\documentclass[10pt,twocolumn,letterpaper]{article}

\usepackage{wacv}
\usepackage{times}
\usepackage{epsfig}
\usepackage{graphicx}
\usepackage{amsmath}
\usepackage{amssymb}

\usepackage{booktabs}       %
\usepackage{subcaption}
\usepackage{graphicx}
\usepackage{amsmath}

\usepackage{enumitem}

\wacvfinalcopy %

\newcommand*{\thead}[1]{\multicolumn{1}{c}{\bfseries #1}}

\begin{document}

\title{Neural Image Decompression:\\Learning to Render Better Image Previews}

\author{
  Shumeet Baluja, Dave Marwood, Nick Johnston, Michele Covell\\
  Google AI\\
  Google, Inc.\\
}
\maketitle

\begin{abstract}

  A rapidly increasing portion of Internet traffic is dominated by
  requests from mobile devices with limited- and metered-bandwidth
  constraints.  To satisfy these requests, it has become standard
  practice for websites to transmit small and extremely compressed
  image previews as part of the initial page-load process.  Recent
  work, based on an adaptive triangulation of the target image, has
  shown the ability to generate thumbnails of full images at extreme
  compression rates: 200 bytes or less with impressive gains (in terms
  of PSNR and SSIM) over both JPEG and WebP standards.  However,
  qualitative assessments and preservation of semantic content can be
  less favorable.  We present a novel method to significantly improve
  the reconstruction quality of the original image with no changes to
  the encoded information.  Our neural-based decoding not only
  achieves higher PSNR and SSIM scores than the original methods, but
  also yields a substantial increase in semantic-level content
  preservation.  In addition, by keeping the same encoding stream, our
  solution is completely inter-operable with the original %
  decoder.  The end result is suitable for a range of small-device
  deployments, as it involves only a single forward-pass through a
  small, scalable network.

\end{abstract}

\section{Introduction}
\label{sec:intro}

Compression of high-quality thumbnails is an active area of
research~\cite{DBLP:journals/corr/abs-1710-09926,toderici2016,agustsson2018,jiang1999image,balle2018variational}
as the demand for image content over connections of all speeds
continues to quickly rise. In addition to the decreased download latency
and bandwidth consumption that is particularly important to the ``next
billion users'' (NBU), reducing the compressed-image size also helps with storage
requirements for the billions of thumbnails needed for rapid
access~\cite{webp,cabral2015,apple}.

Two standard measures of compression quality are PSNR and SSIM~\cite{wang2004image}.
However, at such high-compression rates
(200 bytes per thumbnail image, which is 0.033 bpp for $221 \times 221$ thumbnails),
we have found that these metrics do not adequately reflect subjective
preferences.  Therefore, in addition to using PSNR and SSIM, we
measure how well semantic information, in terms of recognizable
objects and scenes, is preserved.

Similarly, at these extreme-compression rates, JPEG and other standard
approaches do not fare well.
Usually, when extreme compression is
required, it is addressed with domain-specific techniques:
for example, faces~\cite{bryt2008compression}, satellite
imagery~\cite{huang2011satellite}, smooth synthetic
images~\cite{orzan2013diffusion}, or
surveillance~\cite{zhu2015dictionary}. For
non-specialized image-compression, \emph{WebP}~\cite{webp} is a leading
compression format.  
When used on small
images, WebP yields better compression than both JPEG and JPEG2000
standards~\cite{webpdev,webpdev2016}.  %

The fundamental operation of
both WebP and JPEG
is a
subdivision of the image into a set of blocks.  Alternative approaches
have used triangulation~\cite{bougleux2009image,davoine1996fractal,demaret2006image,marwood2018}.
The most recent of these,~\cite{marwood2018}, has shown promising
results on a wide variety of natural images.  Their approach creates
an adaptive Delaunay~\cite{delaunay1934sphere} triangulation of the
target image, based on the underlying entropy of the local pixel
distributions.
The result is a mesh in which a larger number of triangles are devoted
to the complex (high-entropy) regions, while smooth patches of the
image are approximated with fewer triangles.  After transmission, the
decoder renders the triangles by interpolating the vertex colors.

\begin{figure*}
  \rotatebox{90}{\footnotesize ~~~~~~c~~~Network~~~~~~~~~~~~~~~~~~~~~~~~~~~~~b~~~Triangulation~~~~~~~~~~~~~~~~~~~~~~a~~~Original}
  \rotatebox{90}{\footnotesize ~~~~~~~~~~Reconstruction~~~~~~~~~~~~~~~~~~~~~~~~~Reconstruction~~~~~~~~~~~~~~~~~~~~~~~~~Image}  
  ~
  \includegraphics[width=0.18\linewidth]{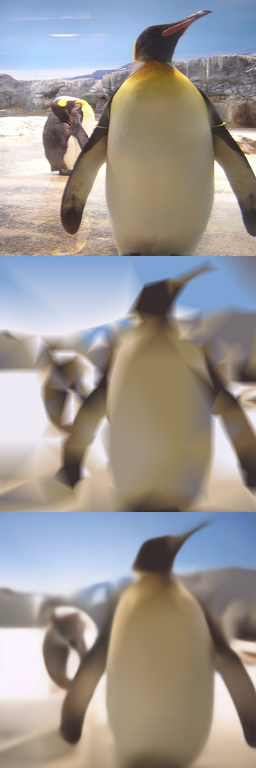}
  \includegraphics[width=0.18\linewidth]{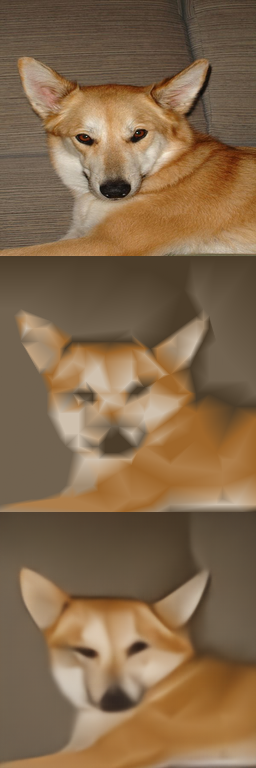}
  \includegraphics[width=0.18\linewidth]{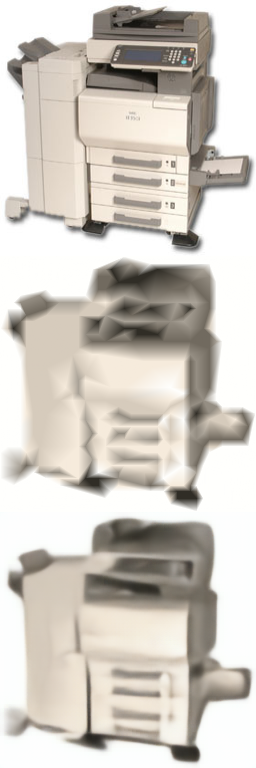}
  \includegraphics[width=0.18\linewidth]{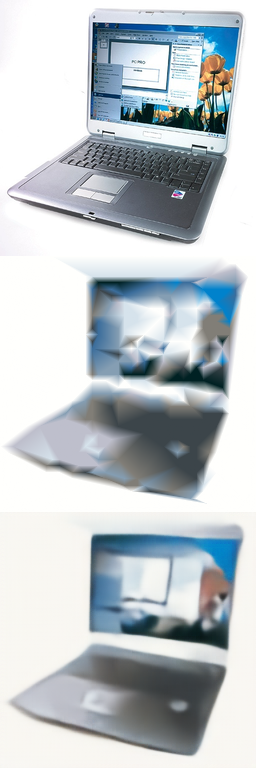}    
  \includegraphics[width=0.18\linewidth]{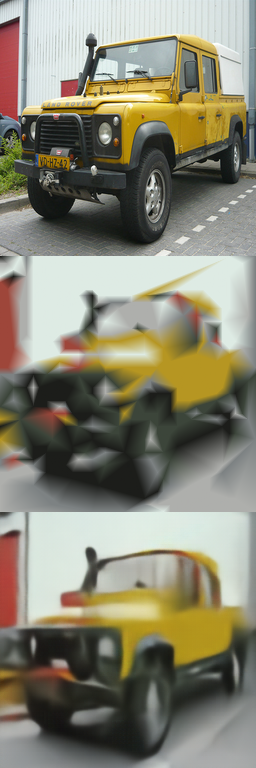}
\caption{Images represented in \textasciitilde 200 bytes.  Five
  original images (row a) are compressed by the
  state-of-the-art triangulation method from~\cite{marwood2018}
  (row b) and by our neural decoder (row c).  A variety of
  qualitative results were intentionally selected --- from good
  (left 2 columns) to poor with many visual artifacts (right 2
  columns).  The middle column is acceptable, though many extraneous
  ``jagged'' artifacts are visible.  See the appendix for more examples.
}

\label{teaser}
\end{figure*}  

The performance of the triangulation method in~\cite{marwood2018}
provides a strong encoder that works well precisely in the regime of
interest: transmission of images under 200 bytes.  At that small size,
image previews can be easily transmitted as part of the original
page-load process on mobile devices or on bandwidth limited
connections~\cite{cabral2015,marwood2018}.  When measured in
terms of %
PSNR or SSIM~\cite{wang2004image},
the triangulation method significantly outperformed JPEG and WebP. However, when the
images were visually inspected,
their visual quality was very uneven:
see Figure~\ref{teaser} (row b) for examples.  Though
some of the images appear very well reconstructed (Figure~\ref{teaser}
left columns), others are unrecognizable when viewed without the
reference.  Other images resulted in spurious edges
formed by the triangulation boundaries
(Figure~\ref{teaser} right columns).  To address these shortcomings,
we replace the decoder with a deep convolutional neural network.  We
ensure that the network remains relatively modest in size for ease of
deployment.
The decoder input-feature
representations played a crucial role for good performance:
we provide details
in Section~\ref{network}.  The results, presented in
Section~\ref{experiments}, reveal not only improved PSNR and SSIM
scores, but also semantic-content preservation that is quantitatively measured as far superior. 

Deep neural networks for compression have been studied in a variety of
configurations, from shallow~\cite{cottrell1988principal,kramer1991nonlinear,jiang1999image}
and deep feed-forward auto-encoders~\cite{balle2018variational,rippel2017,theis2017,balle2017iclr}
to recurrent neural nets/LSTMs for %
variable-length encodings~\cite{toderici2016,toderici2017full}.
Others have taken approaches more closely tied in spirit to ours:
employing established encodings as the inputs and using neural
networks as the basis for a new decoder with improved performance.
These techniques effectively learn a mapping from decompressed patches
back to the original image, for example to remove JPEG compression
artifacts~\cite{yu2016deep,Svodboda2016,cavigelli2017cas}.  Finally,
though we do not explore generative adversarial networks (GANs) in
this paper, we will briefly address how they can easily be used in a
manner similar to other super-resolution and compression
studies~\cite{ledig2016photo,agustsson2018}.

\section{Triangulation of Images: Encoding \& Decoding}
\label{sec:algorithm}

In this section, we review the triangulation approach presented
in~\cite{marwood2018}; this yields a state-of-the-art compressed
encoding that is used (indirectly) as the input for our neural decoder
(presented in the next section).

In~\cite{marwood2018}, the compressed representation of an image
describes a list of colored
vertices and a color table. The vertices lie on a regular grid of size $M
\times M$ and the edges of the grid lie on the edges of the
image. The vertex color is an index into the color table. Their ``triangle-based'' decoder constructs a Delaunay triangulation of the
vertices on a raster image of size $N \times N$ where $M << N$.
Each raster pixel in a triangle is colored using a linear interpolation
of the colors of its triangle's vertices.

Their encoder uses a stochastic-hillclimbing optimizer to find the
vertices and color table that optimize the output of their decoder, i.e.,
that produce %
a good Delaunay triangulation and raster pixel
colors from their decoder algorithm. In this way, the encoder is
optimized specifically for their decoder.

\begin{figure*}
\begin{subfigure}{0.79in}
\includegraphics[width=0.79in]{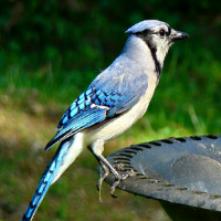}
\vskip -0.05in
\caption*{original ~~~~~~~~~image}
\end{subfigure}~ ~
\begin{subfigure}{0.79in}
\includegraphics[width=0.79in]{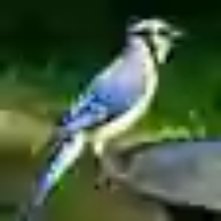}
\vskip -0.05in
\caption* {webp: 370~bytes}
\end{subfigure}
\begin{subfigure}{0.79in}
\includegraphics[width=0.79in]{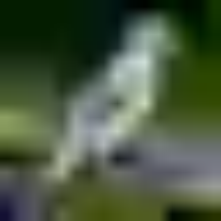}
\vskip -0.05in
\caption* {webp:\\~~~~86~bytes}
\end{subfigure}
~~
\begin{subfigure}{0.79in}
\includegraphics[width=0.79in]{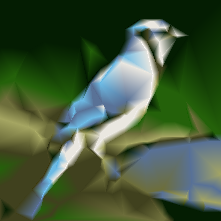}
\vskip -0.05in
\caption* {352~bytes,\\~~~~76~grid}
\end{subfigure}
\begin{subfigure}{0.79in}
\includegraphics[width=0.79in]{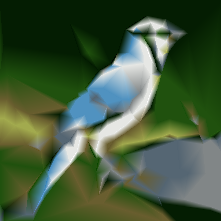}
\vskip -0.05in
\caption* {300~bytes,\\~~~~61~grid}
\end{subfigure}
\begin{subfigure}{0.79in}
\includegraphics[width=0.79in]{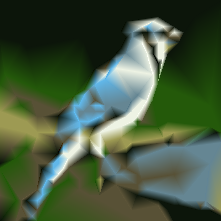}
\vskip -0.05in
\caption* {250~bytes,\\~~~~43~grid}
\end{subfigure}
\begin{subfigure}{0.79in}
\includegraphics[width=0.79in]{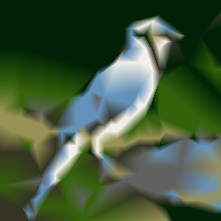}
\vskip -0.05in
\caption* {202~bytes,\\~~~~33~grid}
\end{subfigure}
\begin{subfigure}{0.79in}
\includegraphics[width=0.79in]{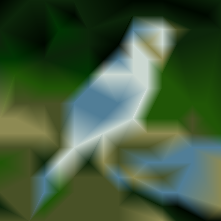}
\vskip -0.05in
\caption* {100~bytes,\\~~~~16~grid}
\end{subfigure}
\hrule
~\\
\begin{subfigure}{0.79in}
\includegraphics[width=0.79in]{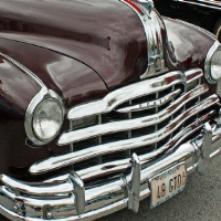}
\vskip -0.05in
\caption* {original ~~~~~~~~~image}
\end{subfigure}~ ~
\begin{subfigure}{0.79in}
\includegraphics[width=0.79in]{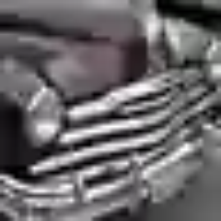}
\vskip -0.05in
\caption* {webp: 438~bytes}
\end{subfigure}
\begin{subfigure}{0.79in}
\includegraphics[width=0.79in]{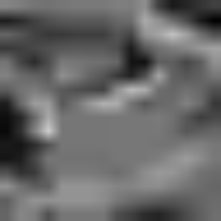}
\vskip -0.05in
\caption* {webp:\\~~~~84~bytes}
\end{subfigure}
~~
\begin{subfigure}{0.79in}
\includegraphics[width=0.79in]{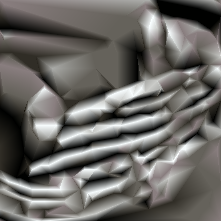}
\vskip -0.05in
\caption* {400~bytes,\\~~~~56~grid}
\end{subfigure}
\begin{subfigure}{0.79in}
\includegraphics[width=0.79in]{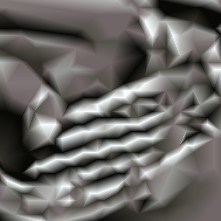}
\vskip -0.05in
\caption* {298~bytes,\\~~~~33~grid}
\end{subfigure}
\begin{subfigure}{0.79in}
\includegraphics[width=0.79in]{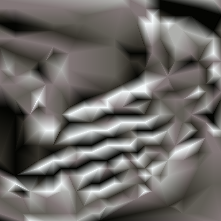}
\vskip -0.05in
\caption* {252~bytes,\\~~~~30~grid}
\end{subfigure}
\begin{subfigure}{0.79in}
\includegraphics[width=0.79in]{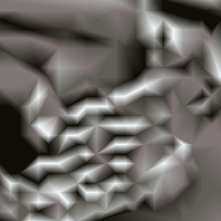}
\vskip -0.05in
\caption* {196~bytes,\\~~~~24~grid}
\end{subfigure}
\begin{subfigure}{0.79in}
\includegraphics[width=0.79in]{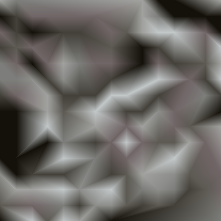}
\vskip -0.05in
\caption* {104~bytes,\\~~~~15~grid}
\end{subfigure}

    \caption{Visualizing the results vs. image byte size.  For each
      of the two original images (left-most column), the image is shown
      after compression with WebP at two levels (roughly 400 \& 100 bytes) and five compression levels
      for the triangulation approach.  The grid-size ($M$) is also
      given. Figure adapted with permission~\cite{marwood2018}. }
    \label{triimages}
    
\end{figure*}
\vspace{0.1in}

We built our decoder to directly operate on the output of the
state-of-the-art encoder presented in~\cite{marwood2018}
because of its good performance %
across a wide variety of natural
images.
and because that encoder has been proposed as a profile in the
next-generation WebP standard~\cite{webp_future}.
By strictly adhering to this as our
input with no modifications, wide deployment becomes substantially
easier.
Ensuring this interoperability of decoders is
an important feature since some very low-end devices may not support
even our light-weight decoding network; therefore, we need to be able
to seamlessly
back-off to~\cite{marwood2018}'s decoder.
For those devices that can support forward propagation through our
simple network, we will demonstrate substantially improved images in
both reconstruction quality and
recognizability.  Unlike other profile-based compression approaches,
this interoperability ensures that the
encoder does not need to know which decoder that the client is using.
In fact, if necessary, a mix of different decoders can all be supported by exactly the
same bit-stream.

Increasing the vertex grid size (making $M$ larger) increases the encoded rate while
reducing distortion. Figure~\ref{triimages} shows sample decompressed
images with grid sizes ranging from $15 \times 15$ to $76 \times 76$
and compressed sizes ranging from 100 to 400 bytes.  In the examples
shown in Figure~\ref{triimages},
the types of errors that the triangle-shading codec introduces
become evident.  As each triangle needs to encode more of the image,
the jagged edges of the triangles introduce spurious features and
misalignments (see the car's front grill in Figure~\ref{triimages}).
Nonetheless, it is interesting to note that even at these extreme
compression levels many colors and much of the shading remain intact.
More examples are presented in Figure~\ref{nnresults} and the appendix; see the ``interpolated'' column.

To provide insight into the actual triangulations computed, see Figures~\ref{overlay}
(right column) and~\ref{inputs} (``edges'' column).  As can be seen,
triangles are more densely concentrated in the high-entropy regions of
the image.  In contrast, the uniform regions of the input image are
adequately represented by fewer triangles. 

\section{Neural Decoding}
\label{network}

\begin{figure*}[h]

  \begin{minipage}{1.5in}
\caption{Two sample triangulations, shown enlarged.  Note the jagged
  edges and discontinuities introduced.  Also note that
  triangles with similar color vertices may yield information about
  shading and color consistency along inferrable directions; this is
  not taken into account when triangles are shaded independently.}
\label {overlay}
  \end{minipage}
  ~~~~~~
  \begin{minipage}{6in}
  \includegraphics[width=0.8\linewidth]{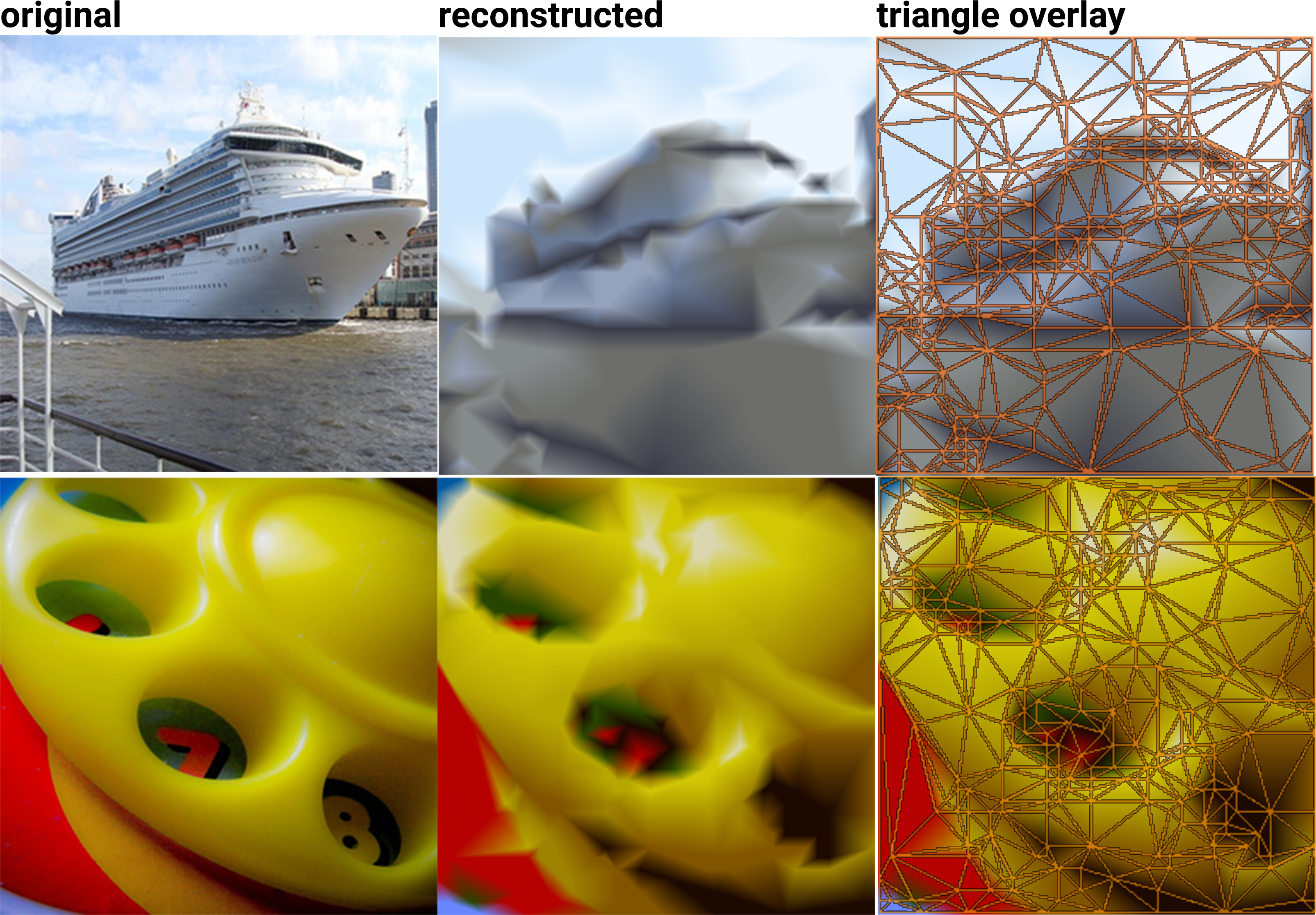}
  \end{minipage}

\end{figure*}

Let us examine a few sample triangulations in detail to see where
there is room for improvement: see Figure~\ref{overlay}.  The most
salient observations are: (1) there are severe jagged edges in the
image (see both images) and (2) discontinuities in straight lines appear (see the boat-deck outline).
These are caused by triangle boundaries.
Recall that each triangle is in-painted
using only the colors of its own vertices.  However, vertices of
nearby triangles have the potential to contain valuable information -
especially when they are assigned the same (or nearly same) color.
For example, in the toy-dial image, notice that many triangles encode
subtle shading differences.  It should be possible to use this
consistency information across triangles in re-rendering the image.

One can imagine a variety of simple techniques to overcome the jagged
edges in the decoded image. %
However, designing the rules to best employ information from close triangles
will likely result in a number of ad-hoc heuristics and thresholds.
Instead, we use a deep neural network to implicitly create the rules
to address both of these shortcomings, based on image statistics.  To
train the network, we start with exactly the same inputs from the
triangulation procedure that were used to render the images shown
above.  For the target output, we use the original image.  
Training proceeds using samples from Imagenet's training set~\cite{deng2014imagenet}.

\subsection {Architecture and Inputs}

A variety of deep convolutional networks have been driving recent
computer-vision research, for example in object detection and
recognition (e.g. the Imagenet challenge~\cite{deng2014imagenet} and
activity recognition~\cite{simonyan2014two}).  For this application,
however, the goal is to take an extremely sparse input and generate a
full image.  We formulate this problem as an \emph{image-translation}
task.  As described by Isola \emph{et. al.}, Image translation is the
task of ``translating one possible representation of a scene into
another, given sufficient training data ... the setting is always the
same: predict pixels from pixels''~\cite{isola2017image}.

Unlike the more common object-identification tasks, where the end
result is a classification, here the result is a full
image. Therefore, it is important to be able to recreate details
from the inputs while allowing for non--spatially-local influences to
direct larger features and impose global consistency.  The need to
have both details from the original image and potentially global
coordination of the generated image has resulted in a variety of
fine$\rightarrow$coarse$\rightarrow$fine architectures such as ``hourglass'' and
``u-net''~\cite{isola2017image, ronneberger2015u}.  These
architectures pass the inputs through a series of convolution layers
that progressively downsample the image.  After the smallest layer is
reached, the process is reversed and the image is expanded back to the
desired size.

\begin{figure*}[t]
\center
  \includegraphics[width=0.9\linewidth]{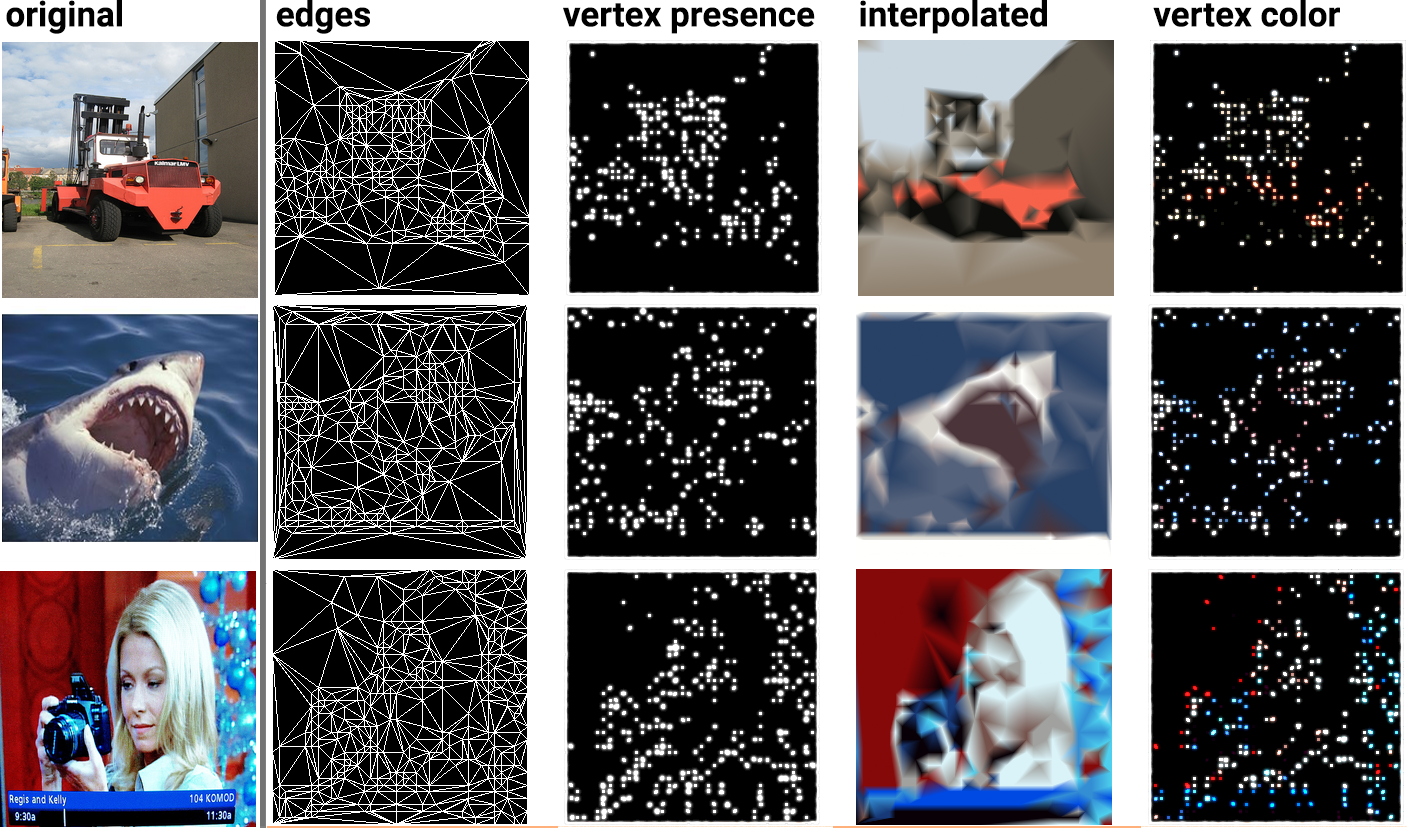}
\caption{Eight input channels.  Left: Original Image (not used as
  input, only shown for reference). Channel 1: Delaunay-edge map
  (binary image).  Channel 2: Binary vertex-location map (visibility
  enhanced for printing).  Channels 3-5: Linear-interpolation
  reconstruction.  Channels 6-8: Colored vertex-location map. (Channel 2
  is also given, in order to correctly represent pure-black vertex colors.)}
\label{inputs}
\end{figure*}  

One of the largest differences between the previous image-to-image
translation work and ours is that our inputs are not the typical
3-channel images.  Instead, they are composed of 8 channels
(Figure~\ref{inputs}): (channel 1) the edge image -
a binary image showing the edges created by the
Delaunay triangulation;  (channel 2) the binary vertex-presence image;
(channels 3-5) the reconstruction using the original system's
bilinear-interpolation approach~\cite{marwood2018}; and
(channels 6-8) the RGB color-vertex image showing
the color assigned to each vertex (with black
everywhere else).%
\footnote{The decision to use images as inputs into the network is not the only possible
approach.  For example, after decoding the transmission, the series of
{vertex+color} tuples could be directly used.  We did not pursue this
avenue since, in addition to learning the image-translation problem,
it would require the network to learn how to triangulate and how to
map between the real-value inputs and coordinates.  Further, more
complex measures would be required to handle the variable number of
vertices.  All of these are avoided by using the eight-channel,
image-like input in which the spatial information is explicitly
maintained and the triangulation's edges directly given.}

Beyond good reconstruction performance, an equally important
consideration for this study was the simplicity/size of the final
decoding network --- keeping computation requirements manageable is
crucial for large-scale device deployment.  An enormous number of
architectures and a variety of approaches were empirically examined.
Because of space limitations, we provide a brief summary of them
here.  We tried architectures ranging from image-translation
(e.g. pix2pix~\cite{isola2017image},
cycle-gan~\cite{zhu2017unpaired}), to shape-encoding/decoding networks
(e.g., where the bottleneck is a set of geometric descriptions), to
progressive-completion networks~\cite{toderici2016,toderici2017full}. The approach that
provided the best trade-off, in terms of reconstruction quality
vs. simplicity, was the stacked hourglass network described below.  The
hourglass network is also simple enough to meet the NBU-application’s
requirements since, in NBU areas, processor computational limitations
are prevalent in the available mobile devices.  As a secondary
benefit, the number of hourglass networks (\emph {e.g. stack size})
can be adjusted according to computational availability, though, as
will be described in the experiments, even a single hourglass provides substantial
benefits.

\begin{figure*}
  \centering
  \includegraphics[width=0.8\linewidth]{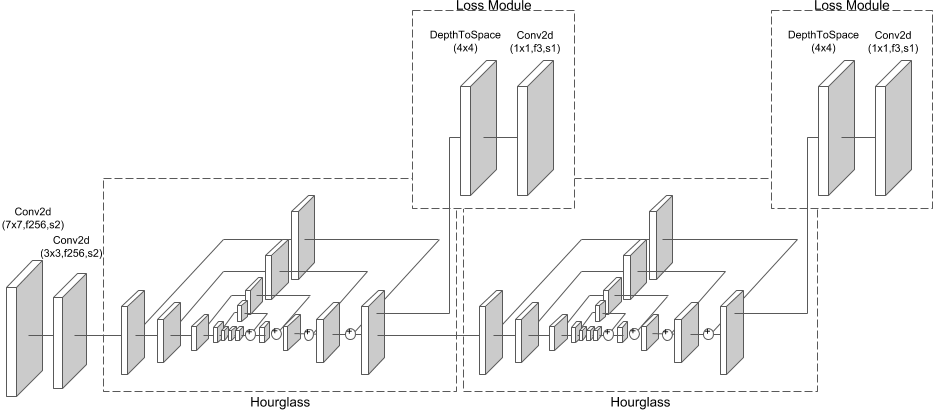}
  \caption{Modified Stacked-Hourglass network from~\cite{newell2016stacked},
    including intermediate loss.}
  \label{netarch-high-level}  
\end{figure*}

The remainder of this study uses the most promising of
these: the hourglass network.  The input images have a resolution of
$256 \times 256$ with 8 channels and a batch size of 32.  The output is
an RGB image of the same resolution.  Our network
(Figure~\ref{netarch-high-level}) is based on the Stacked Hourglass
in~\cite{newell2016stacked}. We apply a
Conv2d(size=7x7,filters=256,stride=2) to the $256 \times 256 \times 8$
input, then a Conv2d(3x3,f256,s2) to bring the dimensions to $64
\times 64 \times 256$. This feeds into an Hourglass as described
in~\cite{newell2016stacked} except 1) when downscaling, each MaxPool
layer is replaced by a
layer that stacks the values of each $2 \times 2$ spatial block
depth-wise (a SpaceToDepth(2x2) layer) followed by a
Conv2d(3x3,f256,s1) and, 2) when upscaling, each
nearest neighbor upsampling is replaced by a DepthToSpace(2x2), the inverse
of a SpaceToDepth, followed by a Conv2d(3x3,f256,s1).
The Hourglass output is added
to the Hourglass input and passed to the next Hourglass. We stacked
two Hourglass networks.

To apply intermediate supervision as described in
~\cite{newell2016stacked}, we split an intermediate Loss Module off the
output of every Hourglass. It is a DepthToSpace(4x4) and a
Conv2d(1x1,f3,s1) with a Tanh activation to get us to a $256 \times 256$ RGB
image. During training, we apply a mean-squared-error loss between
this and the original ground truth image to maximize PSNR.
During inference, the
network's prediction is the $256 \times 256$ RGB image in the second (final)
Hourglass's Loss Module.  Every layer is followed by Batch Norm and
Relu except the final layer (with the Tanh).
We use the Adam Optimizer~\cite{adam2014} with
learning rate of 0.1.

\section{Experimental Results}
\label{experiments}

The network described in the previous section was trained for 2.2 million
steps on five asynchronous GPUs: this was approximately
15~days of continuous training.  Testing was conducted on 20,000~images
drawn from the ImageNet validation set; these were not used
elsewhere in training.

\begin{figure*}

  \begin{minipage}{1.5in}
    \caption {
      Seven results comparing decoding via~\cite{marwood2018} (interp.) and
      our neural network (nn) decoding.  Rows A-E are good quality reconstructions.
      Rows F-G show examples where both methods do poorly.
      Also given with each image pair are the
      PSNR and SSIM scores without (left) and with (right) neural
      network decompression. Additional examples (plus WebP results) can be found in the appendix. \\ ~ \\
      Detailed notes: \\
      A: Note the jagged lines along the window. \\
      B: Both images are recognizable, but our method provides
      cleaner lines. \\
      C: Note the severe artifacts in the vertical lines. \\
      D: Note the smooth shading from the lights and the hair lines. \\
      E: Note the problem with the two thin triangles starting from the upper left corner. \\
      F: Neither method provides a good reconstruction but our decoder gives a slightly better rendition. \\
      G: Compression rate is too extreme for either method to provide
      recognizable results.\\}
    \label{nnresults}
  \end{minipage}\hfill\begin{minipage}{5in}
    \small
    ~~~~~~~~~~~~~~~~~~original~~~~~~~~~~~~~~~~~~~~~~~~~interp.~\cite{marwood2018}~~~~~~~~~~~~~~~~~~~~~~neural (ours)~~~~~~~~~~~~~~~~~~interp. v nn\\
    \begin{minipage}{0.05in} A.
      \end{minipage}\hfill\begin{minipage}{3.7in}
    \includegraphics[width=3.7in]{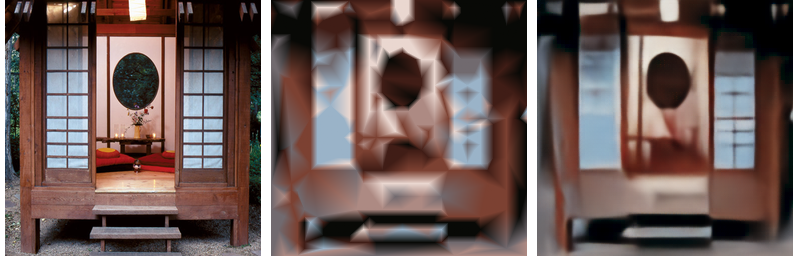}
    \end{minipage}\hfill\begin{minipage}{1.05in} \footnotesize \noindent
      PSNR: 17.74 v \textbf{19.00} \\ SSIM: 0.39 v \textbf{0.46} \\
    \end{minipage} \\
    \begin{minipage}{0.05in} B.
      \end{minipage}\hfill\begin{minipage}{3.7in}
    \includegraphics[width=3.7in]{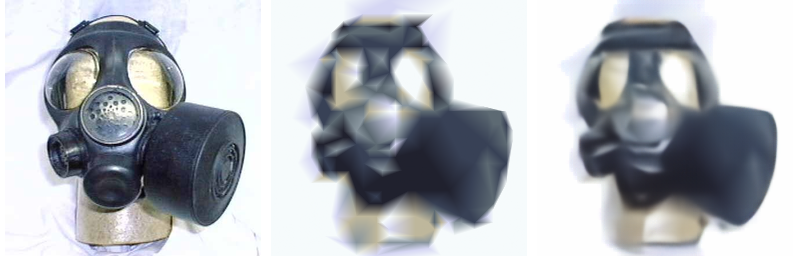}
    \end{minipage}\hfill\begin{minipage}{1.05in} \footnotesize \noindent
      PSNR: 18.01 v \textbf{19.46} \\ SSIM: 0.48 v \textbf{0.54} \\
    \end{minipage} \\
    \begin{minipage}{0.05in} C.
      \end{minipage}\hfill\begin{minipage}{3.7in}
    \includegraphics[width=3.7in]{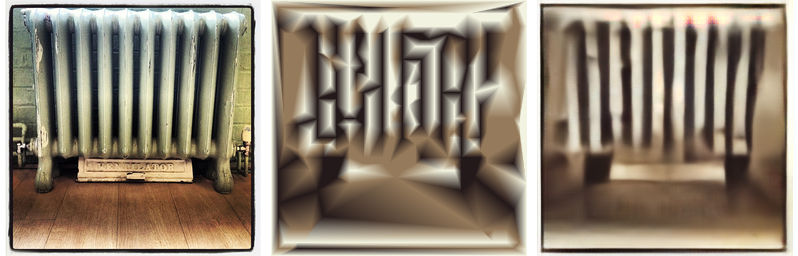}
    \end{minipage}\hfill\begin{minipage}{1.05in} \footnotesize \noindent
      PSNR: 14.67 v \textbf{18.03} \\ SSIM: 0.42 v \textbf{0.48} \\
    \end{minipage} \\
    \begin{minipage}{0.05in} D.
      \end{minipage}\hfill\begin{minipage}{3.7in}
    \includegraphics[width=3.7in]{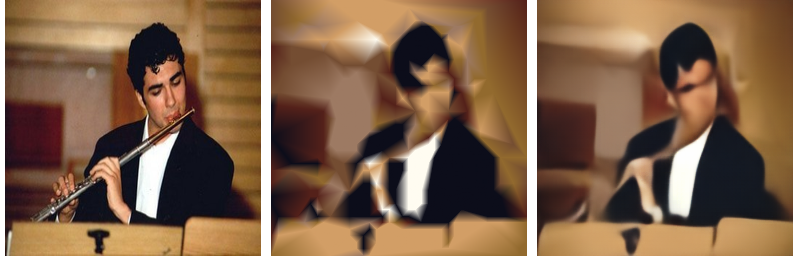}
    \end{minipage}\hfill\begin{minipage}{1.05in} \footnotesize \noindent
      PSNR: 21.54 v \textbf{23.44} \\ SSIM: 0.73 v \textbf{0.78} \\
    \end{minipage} \\
    \begin{minipage}{0.05in} E.
      \end{minipage}\hfill\begin{minipage}{3.7in}
    \includegraphics[width=3.7in]{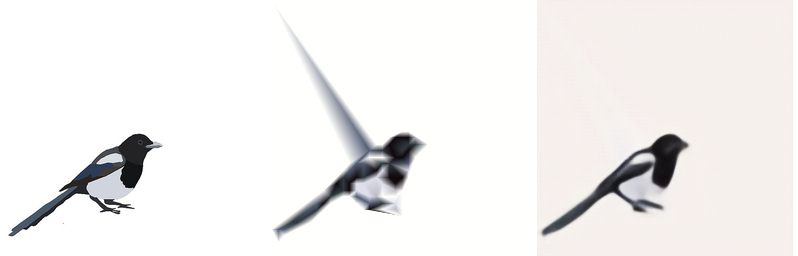}
    \end{minipage}\hfill\begin{minipage}{1.05in} \footnotesize \noindent
      PSNR: 21.47 v \textbf{22.90} \\ SSIM: 0.90 v \textbf{0.96} \\
    \end{minipage} \\
    \begin{minipage}{0.05in} F.
      \end{minipage}\hfill\begin{minipage}{3.7in}
    \includegraphics[width=3.7in]{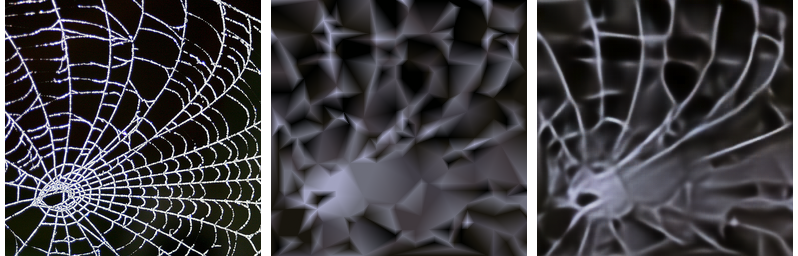}
    \end{minipage}\hfill\begin{minipage}{1.05in} \footnotesize \noindent
      PSNR: 11.66 v \textbf{12.12} \\ SSIM: 0.15 v \textbf{0.24} \\
    \end{minipage} \\
    \begin{minipage}{0.05in} G.
      \end{minipage}\hfill\begin{minipage}{3.7in}
    \includegraphics[width=3.7in]{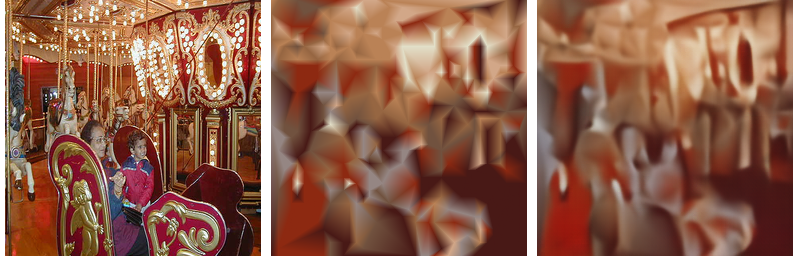}
    \end{minipage}\hfill\begin{minipage}{1.05in} \footnotesize \noindent
      PSNR: 15.86 v \textbf{16.19} \\ SSIM: 0.21 v \textbf{0.24} \\
    \end{minipage} \\
  \end{minipage}
\end{figure*}  

In addition to Figure~\ref{teaser}, Figure~\ref{nnresults} and the appendix
provide more
comparisons to the interpolated images and their respective PSNR and
SSIM (Structural Similarity Index~\cite{wang2004image}) scores.
Overall, when measured on the entire testing set, we are
able to outperform the triangulation approaches in both PSNR and SSIM.

\begin{itemize}[leftmargin=0.2in]
\item \emph {For PSNR:} Triangulation scored 20.7 dB and our neural approach scored
21.7 dB. In this range, a 1-dB PSNR increase is extremely valuable.  Out of
the 13,000 examples examined, 12,810 (98.5\%) showed improved PSNR via the
neural decoding.  Comparing the two approaches using a standard
$t$-test on the PSNR,  $p < 0.0001$.

\item \emph {For SSIM}: Triangulation scored 0.51 and our neural approach scored
0.54.  Out of the 13,000 examples examined, 12,255 (94.2\%) were
improved via the neural decoding. Comparing the two approaches using a
standard $t$-test on SSIM, $p < 0.0001$.
\end{itemize}

Importantly, recall that the triangulation method~\cite{marwood2018} also outperformed
JPEG and WebP, which, in turn, equals or
outperforms JPEG2000~\cite{webpdev,webpdev2016}.

To better understand what the network was encoding, an extensive grid
search was also performed to determine which channels were actually
necessary. For space reasons, we cannot recreate all of the results
here.  A few salient findings, however, are worth noting: (1) The best
performing network was one that received all the eight channels as input;
(2) If we removed the interpolated image (as created
by~\cite{marwood2018}) from the inputs, the PSNR performance drops
approximately 0.75 dB; (3) Interestingly, if we used \emph{only} the
interpolated as input, the PSNR performance drops 0.5 dB;
(4) Finally, while we used 2 stacked Hourglasses in this work, the
results with 1 Hourglass or 3 stacked Hourglasses were almost
identical; any variation was likely due to the stochasticity in the
training procedure. On a mobile or computationally constrained device,
a single Hourglass can be used.  This decision on the complexity of
the decoder can be made on a pre-device basis and all will work \emph{on
  exactly the same encoding}.

\subsection {Semantic Content Preservation}

The quantitative results, in terms of PSNR and SSIM, reveal a
significant improvement for extremely compressed images.  Despite the
numeric improvements, however, it is important to assess whether the
images are qualitatively better.  The simplest,
though resource-consuming, method is to employ human raters.  We propose a novel technique using
a well-trained classification network as an automated proxy.

\begin{table*}
  \caption{Semantic Similarity: Comparing  Classification Vectors of Original and Compressed Images}
  \vspace{0.1in}
  
  \centering
  \begin{tabular}{r||c|ccc}
                 &  \emph{(Lower Better)} & & \emph{(Higher Better)} &\\    
    &  $L_2$ Error & Recall Top-1 & Recall Top-5 & Recall Top-10\\
    \midrule\midrule    
    interpolated~\cite{marwood2018}       & 39.5          & 0.05 & 0.13 & 0.15 \\
    nn-Decoded  (our method)      & \textbf{36.0} & \textbf{0.17} & \textbf{0.33} & \textbf{0.38} \\
    \midrule
    interpolated+Blur x1                  & 39.5          & 0.11 & 0.26 & 0.30\\
    interpolated+Blur x5                  & 43.0          & 0.08 & 0.18 & 0.22\\
    \end{tabular}
\label{semantic}
\vspace{-0.1in}
\end{table*}      

For our experiments, we employ a pre-trained state-of-the-art
classifier, Inception ResNet v2 (IR2), which produces a 1000-dimension
``classification vector'' prior to the final soft-max layer
representing the classification of the objects in the image.  On the
ImageNet challenge, IR2 has has a top-1 single-crop error rate of
19.9\% on the 50,000 image validation set, and a top-5 error rate of
4.9\% \cite{szegedy2016inception}. For each of our test images, we use
the original image, the interpolated image and the neural decoded
image: $t^{\text{orig}}$, $t^{\text{interp}}$, and $t^{\text{nnDec}}$.
Passing each of these through IR2 produces classification vectors
$c^{\text{orig}}$, $c^{\text{interp}}$, and $c^{\text{nnDec}}$. We measure
the similarity between $c^{\text{orig}}$ and $c^{\text{interp}}$ and
between $c^{\text{orig}}$ and $c^{\text{nnDec}}$ in two ways: (1) the $L_2$
difference between the classification vectors, and (2) whether the top
classification in $c^{\text{orig}}$ appears in the top-1,~-5, and~-10
positions of $c^{\text{interp}}$ and $c^{\text{nnDec}}$.

Note that this is \emph{not} equivalent to checking the ground-truth
classification.  The goal of our compression task is not to alter the
original image to make a wrong classification correct, it is to
achieve the \emph{same} classification as the original.  Finally, we
remark that with such aggressive compression rates, we do not expect
all images to be recognized; for example, images in which the object
of interest does not cover a large portion of the image, the object
may be lost.  Nonetheless, for images in which the object is large,
these metrics elucidate how recognizable the object remains.

The results are presented in the first two rows of
Table~\ref{semantic}.  There is more than a 10\% decrease in the $L_2$
error using the neural-network decoding.  However, the largest
benefit comes when looking at the recall measures.  Looking at recall
in the top-1 position, the results are {\bf 300\% improved (3$\times$)} and at
top-10, they remain approximately 2.5$\times$ improved.  This large
gain indicates that the content of the image remains far more
recognizable using our neural-network decoder.

Upon first glance at our decoded neural-network images, it is tempting
to wonder if much of the recognition improvement is coming from simply
blurring the triangulated image.  Though we would not expect an
improvement in PSNR or SSIM from additional blurring, it is possible
that the Inception-Resnet-V2 network is not robust to the types of
edges seen in Figure~\ref{nnresults}.  We explicitly checked that
possibility, to ensure that the network is not acting as an overly
complex approach to a simple blur operation.  Instead of decoding with
a neural network, we use the method from~\cite{marwood2018} followed
by Gaussian blurring ($r$=2).  We create two new test sets: the first
with a single pass of a blur filter and the second with 5 sequential
passes.  The last two rows of Table~\ref{semantic} show the
performance after the added blurring. The results, though improved, do
not match those our NN-based decoding.  And, as expected, the PSNR and
SSIM rates decline for both sets over the base triangulation results
reported above (PSNR: blur1: 20.6, blur5: 19.6, SSIM: blur1: 0.50,
blur5: 0.46).  Visually, it appears that the neural approach is
smoothing the harsh color transitions created by the triangulation.
However, based on the PSNR/SSIM scores and the similarity of the
classification vectors, the neural network's effect is well targeted:
the edges and details required to maintain the object identity and
similarity to the original image are preserved.

\section {Discussion \& Future Work}

The application of neural networks to image decompression is not only
of interest to researchers and practitioners, as witnessed by the vast
amount of neural image compression literature, but also will have a large and
socially important impact: allowing efficient discovery/browsing of
visual content for the ``next-billion users" whose bandwidth is limited
and expensive.  We have found that the impact of using neural
networks in place of the current triangle-shading decoder results in
consistent and very significant quantitative and qualitative
improvements to the final image quality.

By casting the task of decompression into an image-to-image--translation
problem, we were able to generate images that, when
compared to recently released state-of-the-art compression techniques,
more closely resemble the original image in terms of the standard
quantitative metrics such as PSNR and SSIM.  More importantly,
they far exceeded the previous method~\cite{marwood2018} in preserving \emph{semantic}
quality.  The results come in an operating regime of extreme
compression where there is large practical interest, but existing
compression schemes do not fare well.

These improved results are somewhat surprising since, on each
encoding, the encoder is explicitly optimizing for the best results
from the triangle-based decoder in~\cite{marwood2018}.
Yet, we
are able to provide better reconstructions with neural decompression
\textit{without changing the encoder at all}. This points the way for
efforts to replace full H.264 decoders with neural approaches without
changing the already deployed video encoders.

This study leads to many avenues of future work.  First,
simultaneously to the development of this study, Generative
Adversarial Networks were in parallel developed for
compression~\cite{agustsson2018}.  Beyond using GANs for error
signals, they also make clever use of the ability for GANs to
synthesize, rather than compress.  Though they operate on larger
images at higher bit rates, many of the same approaches, including
using GANs to augment the objective functions, can easily be
incorporated.

Second, although not discussed in this paper, an interesting side
finding was early evidence that it is possible to
train a network to infer a Delaunay triangulation given just the
vertex points.  In preliminary studies, the network fared much better
than expected in not only finding the same connections, but also in
creating relatively straight edges between the vertices (the output
was a $256\times256$ image).  If these results hold true, this has
potentially broad applicability as the operation of triangulation could
then be integrated into a fully differentiable system.

Third, we should consider that if we know that semantic recall, as
measured by IR2, is important, should it be included as an extra
error term during training?  The answer may not be straightforward --
if it is used, it is possible that the examples generated will take
advantage of small inconsistencies in the training, %
in
the same way that adversarial attacks are remarkably plentiful and
easy to find.  On the other hand, if we train and test on distinct semantic models,
perhaps the semantic recall
will improve without falling into model-specific traps.

\bibliographystyle{ieee}
\bibliography{triangle}

\begin{thebibliography}{10}\itemsep=-1pt

\bibitem{agustsson2018}
E.~Agustsson, M.~Tschannen, F.~Mentzer, R.~Timofte, and L.~V. Gool.
\newblock Generative adversarial networks for extreme learned image
  compression.
\newblock {\em CoRR}, abs/1804.02958, 2018.

\bibitem{balle2017iclr}
J.~Ball\'{e}, V.~Laparra, and E.~P. Simoncelli.
\newblock End-to-end optimized image compression.
\newblock In {\em Int'l. Conf. on Learning Representations (ICLR2017)}, Toulon,
  France, April 2017.
\newblock Available at http://arxiv.org/abs/1611.01704.

\bibitem{balle2018variational}
J.~Ball{\'e}, D.~Minnen, S.~Singh, S.~J. Hwang, and N.~Johnston.
\newblock Variational image compression with a scale hyperprior.
\newblock {\em arXiv preprint arXiv:1802.01436}, 2018.

\bibitem{bougleux2009image}
S.~Bougleux, G.~Peyr{\'e}, and L.~D. Cohen.
\newblock Image compression with anisotropic triangulations.
\newblock In {\em Computer Vision, 2009 IEEE 12th International Conference on},
  pages 2343--2348. IEEE, 2009.

\bibitem{bryt2008compression}
O.~Bryt and M.~Elad.
\newblock Compression of facial images using the {K-SVD} algorithm.
\newblock {\em Journal of Visual Communication and Image Representation},
  19(4):270--282, 2008.

\bibitem{cabral2015}
B.~Cabral and E.~Kandrot.
\newblock The technology behind preview photos.
\newblock https://code.facebook.com/
  posts/991252547593574/the-technology-behind-preview-photos/, 2015.

\bibitem{cavigelli2017cas}
L.~Cavigelli, P.~Hager, and L.~Benini.
\newblock Cas-cnn: A deep convolutional neural network for image compression
  artifact suppression.
\newblock In {\em Neural Networks (IJCNN), 2017 International Joint Conference
  on}, pages 752--759. IEEE, 2017.

\bibitem{cottrell1988principal}
G.~W. Cottrell and P.~Munro.
\newblock Principal components analysis of images via back propagation.
\newblock In {\em Visual Communications and Image Processing'88: Third in a
  Series}, volume 1001, pages 1070--1078. International Society for Optics and
  Photonics, 1988.

\bibitem{davoine1996fractal}
F.~Davoine, M.~Antonini, J.-M. Chassery, and M.~Barlaud.
\newblock Fractal image compression based on delaunay triangulation and vector
  quantization.
\newblock {\em IEEE Transactions on Image Processing}, 5(2):338--346, 1996.

\bibitem{delaunay1934sphere}
B.~Delaunay.
\newblock Sur la sphere vide.
\newblock {\em Izv. Akad. Nauk SSSR, Otdelenie Matematicheskii i Estestvennyka
  Nauk}, 7(793-800):1--2, 1934.

\bibitem{demaret2006image}
L.~Demaret, N.~Dyn, and A.~Iske.
\newblock Image compression by linear splines over adaptive triangulations.
\newblock {\em Signal Processing}, 86(7):1604--1616, 2006.

\bibitem{webpdev}
G.~Developers.
\newblock Comparative study of webp, jpeg and jpeg 2000.
\newblock https://developers.google.com/speed/webp/docs/c\_study, 2010.

\bibitem{webp}
G.~Developers.
\newblock A new image format for the web.
\newblock https://developers.google.com/speed/webp/, 2016.

\bibitem{webpdev2016}
G.~Developers.
\newblock Webp compression study.
\newblock https://developers.google.com/speed/webp/docs/webp\_study, 2016.

\bibitem{huang2011satellite}
B.~Huang.
\newblock {\em Satellite data compression}.
\newblock Springer Science \& Business Media, 2011.

\bibitem{apple}
A.~Inc.
\newblock Using {HEIF} or {HEVC} media on {Apple} devices.
\newblock https://support.apple.com/en-us/HT207022, 2017.

\bibitem{isola2017image}
P.~Isola, J.-Y. Zhu, T.~Zhou, and A.~A. Efros.
\newblock Image-to-image translation with conditional adversarial networks.
\newblock {\em arXiv preprint}, 2017.

\bibitem{jiang1999image}
J.~Jiang.
\newblock Image compression with neural networks--a survey.
\newblock {\em Signal Processing: Image Communication}, 14(9):737--760, 1999.

\bibitem{adam2014}
D.~P. Kingma and J.~Ba.
\newblock Adam: {A} method for stochastic optimization.
\newblock April 2014.
\newblock Available at http://arxiv.org/abs/1412.6980.

\bibitem{kramer1991nonlinear}
M.~A. Kramer.
\newblock Nonlinear principal component analysis using autoassociative neural
  networks.
\newblock {\em AIChE journal}, 37(2):233--243, 1991.

\bibitem{ledig2016photo}
C.~Ledig, L.~Theis, F.~Husz{\'a}r, J.~Caballero, A.~Cunningham, A.~Acosta,
  A.~Aitken, A.~Tejani, J.~Totz, Z.~Wang, et~al.
\newblock Photo-realistic single image super-resolution using a generative
  adversarial network.
\newblock {\em arXiv preprint}, 2016.

\bibitem{marwood2018}
D.~Marwood, P.~Massimino, M.~Covell, and S.~Baluja.
\newblock Representing images in 200 bytes: Compression via triangulation.
\newblock In {\em International Conference On Image Processing}, 2018
  (arxiv:1809.02257).

\bibitem{webp_future}
P.~Massimino.
\newblock Re: Questions about plans for the triangle encoder going into
  next-gen {W}eb{P}.
\newblock e-mail communication, 2018.

\bibitem{newell2016stacked}
A.~Newell, K.~Yang, and J.~Deng.
\newblock Stacked hourglass networks for human pose estimation.
\newblock In {\em European Conference on Computer Vision}, pages 483--499.
  Springer, 2016.

\bibitem{orzan2013diffusion}
A.~Orzan, A.~Bousseau, P.~Barla, H.~Winnem{\"o}ller, J.~Thollot, and
  D.~Salesin.
\newblock Diffusion curves: a vector representation for smooth-shaded images.
\newblock {\em Communications of the ACM}, 56(7):101--108, 2013.

\bibitem{rippel2017}
O.~Rippel and L.~Bourdev.
\newblock Real-time adaptive image compression.
\newblock In {\em International Conference on Machine Learning (ICML), 2017}.
  IEEE, 2017.

\bibitem{ronneberger2015u}
O.~Ronneberger, P.~Fischer, and T.~Brox.
\newblock U-net: Convolutional networks for biomedical image segmentation.
\newblock In {\em International Conference on Medical image computing and
  computer-assisted intervention}, pages 234--241. Springer, 2015.

\bibitem{deng2014imagenet}
O.~Russakovsky, J.~Deng, H.~Su, J.~Krause, S.~Satheesh, S.~Ma, Z.~Huang,
  A.~Karpathy, A.~Khosla, M.~S. Bernstein, A.~C. Berg, and F.~Li.
\newblock Imagenet large scale visual recognition challenge.
\newblock {\em CoRR}, abs/1409.0575, 2014.

\bibitem{simonyan2014two}
K.~Simonyan and A.~Zisserman.
\newblock Two-stream convolutional networks for action recognition in videos.
\newblock In {\em Advances in neural information processing systems}, pages
  568--576, 2014.

\bibitem{Svodboda2016}
P.~Svoboda, M.~Hradis, D.~Barina, and P.~Zemc{\'{\i}}k.
\newblock Compression artifacts removal using convolutional neural networks.
\newblock {\em CoRR}, abs/1605.00366, 2016.

\bibitem{szegedy2016inception}
C.~Szegedy, S.~Ioffe, V.~Vanhoucke, and A.~Alemi.
\newblock Inception-v4, inception-resnet and the impact of residual connections
  on learning.
\newblock {\em arXiv preprint arXiv:1602.07261}, 2016.

\bibitem{theis2017}
L.~Theis, W.~Shi, A.~Cunningham, and F.~Huszár.
\newblock Lossy image compression with compressive autoencoders.
\newblock In {\em International Conference on Learning Representations}, 2017.

\bibitem{toderici2016}
G.~Toderici, S.~M. O'Malley, S.~J. Hwang, D.~Vincent, D.~Minnen, S.~Baluja,
  M.~Covell, and R.~Sukthankar.
\newblock Variable rate image compression with recurrent neural networks.
\newblock {\em ICLR}, 2016.

\bibitem{toderici2017full}
G.~Toderici, D.~Vincent, N.~Johnston, S.~J. Hwang, D.~Minnen, J.~Shor, and
  M.~Covell.
\newblock Full resolution image compression with recurrent neural networks.
\newblock In {\em Computer Vision and Pattern Recognition (CVPR), 2017 IEEE
  Conference on}, pages 5435--5443. IEEE, 2017.

\bibitem{wang2004image}
Z.~Wang, A.~C. Bovik, H.~R. Sheikh, and E.~P. Simoncelli.
\newblock Image quality assessment: from error visibility to structural
  similarity.
\newblock {\em IEEE transactions on image processing}, 13(4):600--612, 2004.

\bibitem{DBLP:journals/corr/abs-1710-09926}
Y.~Watkins, M.~Sayeh, O.~Iaroshenko, and G.~Kenyon.
\newblock Image compression: Sparse coding vs. bottleneck autoencoders.
\newblock {\em CoRR}, abs/1710.09926, 2017.

\bibitem{yu2016deep}
K.~Yu, C.~Dong, C.~C. Loy, and X.~Tang.
\newblock Deep convolution networks for compression artifacts reduction.
\newblock {\em arXiv preprint arXiv:1608.02778}, 2016.

\bibitem{zhu2017unpaired}
J.-Y. Zhu, T.~Park, P.~Isola, and A.~A. Efros.
\newblock Unpaired image-to-image translation using cycle-consistent
  adversarial networks.
\newblock {\em arXiv preprint}, 2017.

\bibitem{zhu2015dictionary}
J.-Y. Zhu, Z.-Y. Wang, R.~Zhong, and S.-M. Qu.
\newblock Dictionary based surveillance image compression.
\newblock {\em Journal of Visual Communication and Image Representation},
  31:225--230, 2015.

\end{thebibliography}
  \vspace{0.3in}

\appendix
\section {Additional Examples}

We provide additional examples of the neural-decoding method's best
and worst performance on both PSNR and SSIM.  Images are from the
ImageNet test set, reported in the paper.

Ten examples of each have been provided in the tables below along with
their metrics and a comparison to the bi-linear--interpolated, non--neural-network
approach and WebP. Because WebP could not target the same rates
on $256 \times 256$ resolution images, the input images were resized to 4 or more times
smaller in each dimension, compressed with WebP, decompressed, and then upscaled
to bring it back to source resolution.

\begin{figure*}[h]
  \vspace*{-0.1in}
  \caption{Best PSNR}
  \vspace*{-0.1in}
  \centering
  \begin{tabular}{cccc}
  \thead{Original Image} & \thead{Interpolated} & \thead{Neural Decoded} & \thead{WebP} \\
\includegraphics[width=0.22\textwidth]{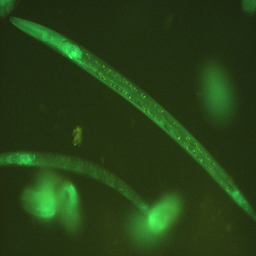} &\includegraphics[width=0.22\textwidth]{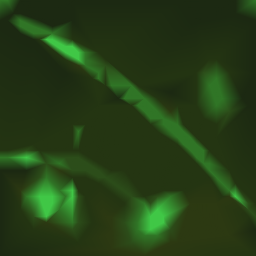} &\includegraphics[width=0.22\textwidth]{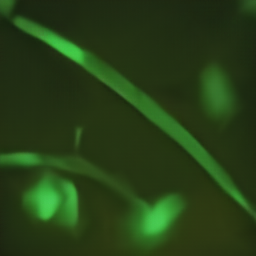} &\includegraphics[width=0.22\textwidth]{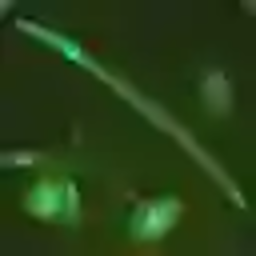} \\

\footnotesize{Ground Truth} & \footnotesize{33.0163 PSNR, 0.9023 SSIM} & \footnotesize{\textbf{34.2382 PSNR, 0.9074 SSIM}}& \footnotesize{28.8307 PSNR, 0.8653 SSIM} \\

\includegraphics[width=0.22\textwidth]{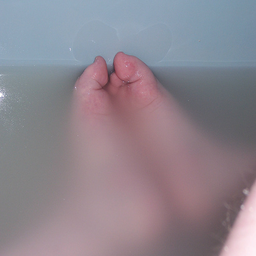} &\includegraphics[width=0.22\textwidth]{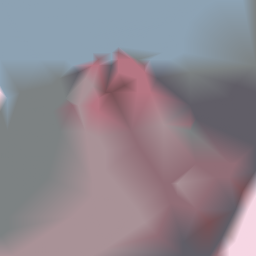} &\includegraphics[width=0.22\textwidth]{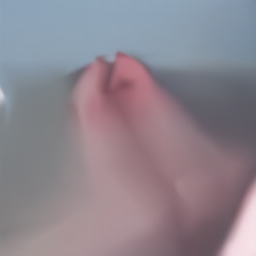}&\includegraphics[width=0.22\textwidth]{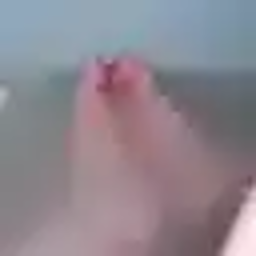} \\
\footnotesize{Ground Truth} & \footnotesize{32.3473 PSNR, 0.9331 SSIM} & \footnotesize{\textbf{34.7422 PSNR, 0.9415 SSIM}} & \footnotesize{33.7351 PSNR, 0.9364 SSIM} \\

\includegraphics[width=0.22\textwidth]{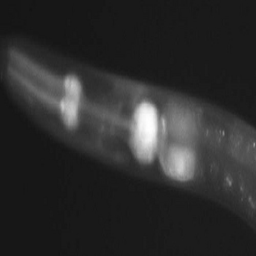} &\includegraphics[width=0.22\textwidth]{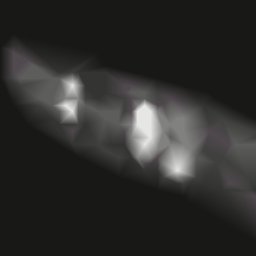} &\includegraphics[width=0.22\textwidth]{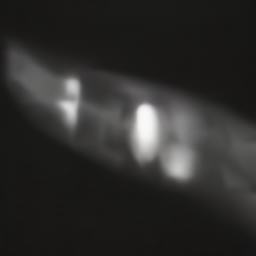}&\includegraphics[width=0.22\textwidth]{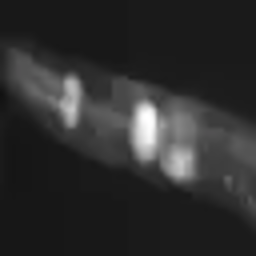} \\
\footnotesize{Ground Truth} & \footnotesize{32.7623 PSNR, 0.9441 SSIM} & \footnotesize{34.8865 PSNR, 0.9535 SSIM} & \footnotesize{\textbf{37.6550 PSNR, 0.9625 SSIM}} \\

\includegraphics[width=0.22\textwidth]{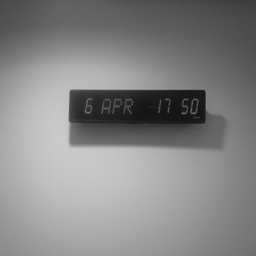} &\includegraphics[width=0.22\textwidth]{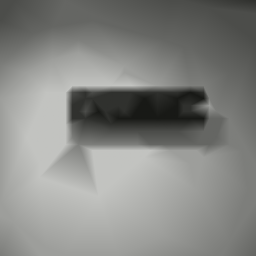} &\includegraphics[width=0.22\textwidth]{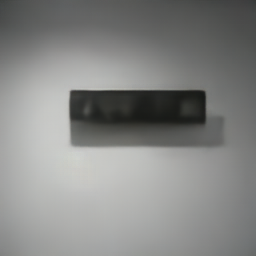} &\includegraphics[width=0.22\textwidth]{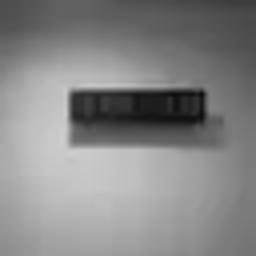} \\
\footnotesize{Ground Truth} & \footnotesize{29.7310 PSNR, 0.9296 SSIM} & \footnotesize{\textbf{35.3078 PSNR, 0.9522 SSIM}} & \footnotesize{33.9794 PSNR, 0.9413 SSIM} \\

\includegraphics[width=0.22\textwidth]{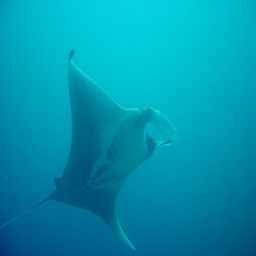} &\includegraphics[width=0.22\textwidth]{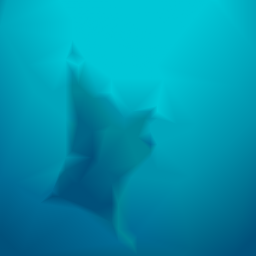} &\includegraphics[width=0.22\textwidth]{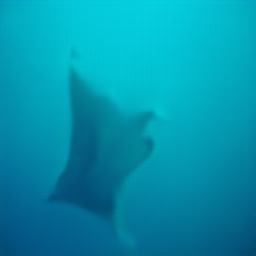} &\includegraphics[width=0.22\textwidth]{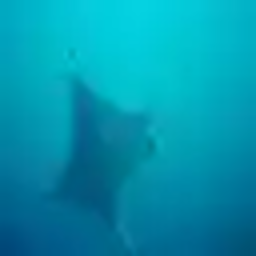} \\
\footnotesize{Ground Truth} & \footnotesize{36.9100 PSNR, \textbf{0.9358 SSIM}} & \footnotesize{\textbf{37.3513 PSNR}, 0.8361 SSIM} & \footnotesize{36.0570 PSNR, 0.8978 SSIM} \\

  \end{tabular}
\end{figure*}

\begin{figure*}[h]
  \vspace*{-0.1in}
  \caption{Best PSNR}
  \vspace*{-0.1in}
  \centering
  \begin{tabular}{cccc}
  \thead{Original Image} & \thead{Interpolated} & \thead{Neural Decoded} & \thead{WebP} \\
\includegraphics[width=0.22\textwidth]{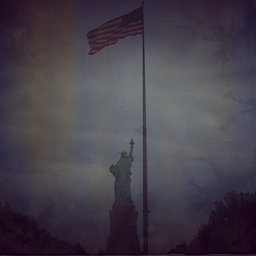} &\includegraphics[width=0.22\textwidth]{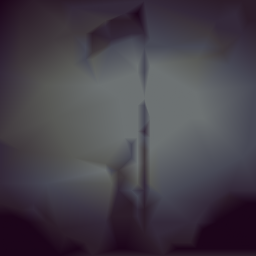} &\includegraphics[width=0.22\textwidth]{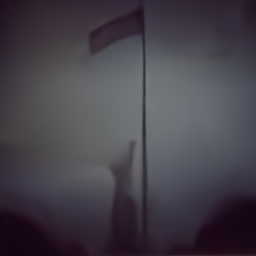} &\includegraphics[width=0.22\textwidth]{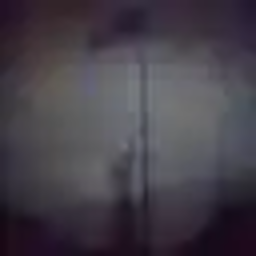}\\

\footnotesize{Ground Truth} & \footnotesize{33.5368 PSNR, 0.9082 SSIM} & \footnotesize{\textbf{34.4789 PSNR, 0.9242 SSIM}}& \footnotesize{33.0402 PSNR, 0.9024 SSIM} \\

\includegraphics[width=0.22\textwidth]{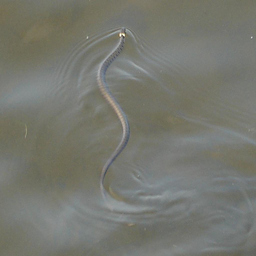} &\includegraphics[width=0.22\textwidth]{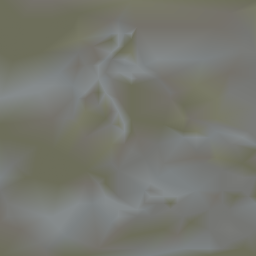} &\includegraphics[width=0.22\textwidth]{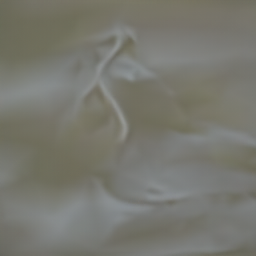} &\includegraphics[width=0.22\textwidth]{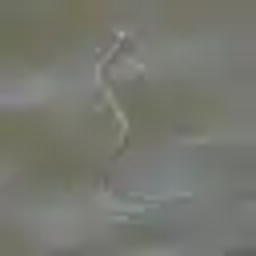}\\
\footnotesize{Ground Truth} & \footnotesize{34.2424 PSNR, 0.8959 SSIM} & \footnotesize{\textbf{34.8164 PSNR, 0.9018 SSIM}} & \footnotesize{33.2535 PSNR, 0.8920 SSIM} \\

\includegraphics[width=0.22\textwidth]{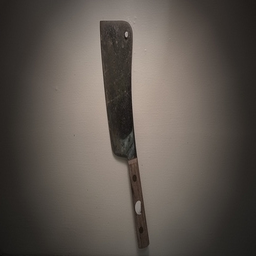} &\includegraphics[width=0.22\textwidth]{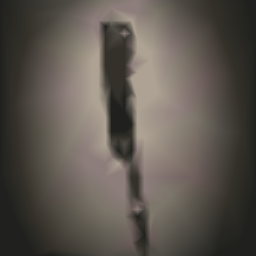} &\includegraphics[width=0.22\textwidth]{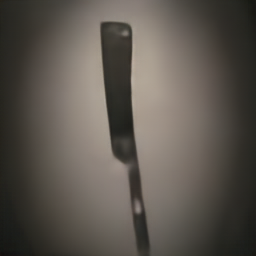} &\includegraphics[width=0.22\textwidth]{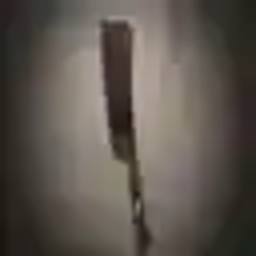}\\
\footnotesize{Ground Truth} & \footnotesize{30.6643 PSNR, 0.9273 SSIM} & \footnotesize{\textbf{34.9344 PSNR, 0.9540 SSIM}} & \footnotesize{31.6522 PSNR, 0.9306 SSIM} \\

\includegraphics[width=0.22\textwidth]{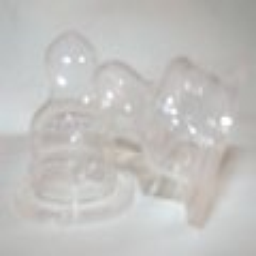} &\includegraphics[width=0.22\textwidth]{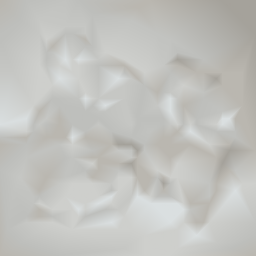} &\includegraphics[width=0.22\textwidth]{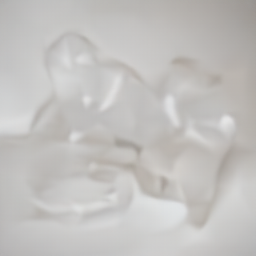} &\includegraphics[width=0.22\textwidth]{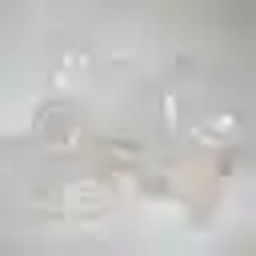}\\
\footnotesize{Ground Truth} & \footnotesize{34.7180 PSNR, 0.9317 SSIM} & \footnotesize{\textbf{35.7485 PSNR, 0.9376 SSIM}} & \footnotesize{33.5282 PSNR, 0.9195 SSIM} \\

\includegraphics[width=0.22\textwidth]{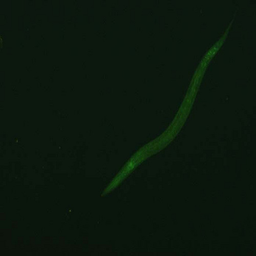} &\includegraphics[width=0.22\textwidth]{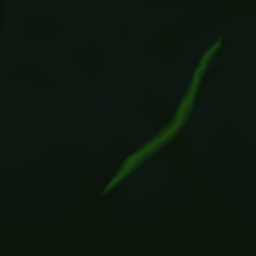} &\includegraphics[width=0.22\textwidth]{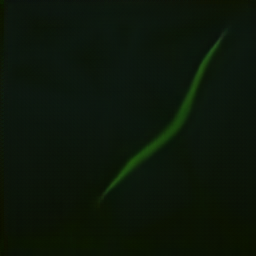} &\includegraphics[width=0.22\textwidth]{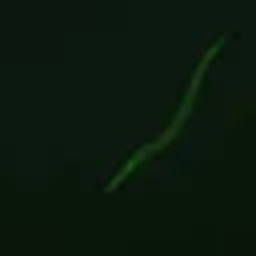}\\
\footnotesize{Ground Truth} & \footnotesize{41.7112 PSNR, \textbf{0.9772 SSIM}} & \footnotesize{37.4974 PSNR, 0.9389 SSIM} & \footnotesize{\textbf{41.7442 PSNR}, 0.9762 SSIM}\\

  \end{tabular}
\end{figure*}

\begin{figure*}[h]
  \vspace*{-0.1in}
  \caption{Worst PSNR}
  \vspace*{-0.1in}
  \centering
  \begin{tabular}{cccc}
  \thead{Original Image} & \thead{Interpolated} & \thead{Neural Decoded} & \thead{WebP} \\
\includegraphics[width=0.22\textwidth]{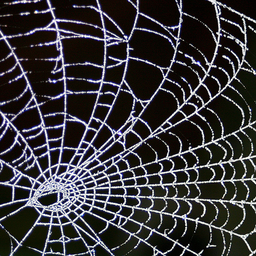} &\includegraphics[width=0.22\textwidth]{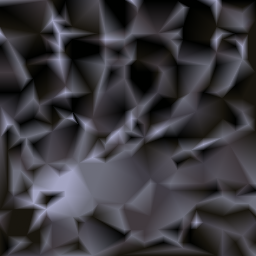} &\includegraphics[width=0.22\textwidth]{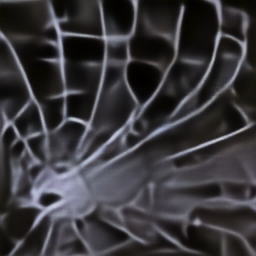} &\includegraphics[width=0.22\textwidth]{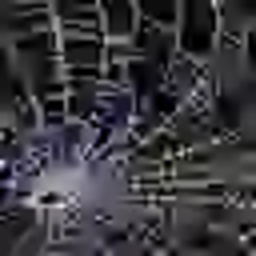} \\

\footnotesize{Ground Truth} & \footnotesize{11.6645 PSNR, 0.1480 SSIM} & \footnotesize{\textbf{12.1248 PSNR, 0.2400 SSIM}} & \footnotesize{11.8997 PSNR, 0.1747 SSIM} \\

\includegraphics[width=0.22\textwidth]{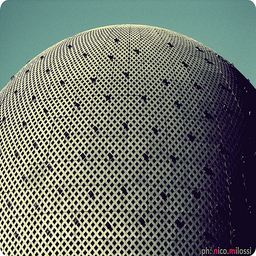} &\includegraphics[width=0.22\textwidth]{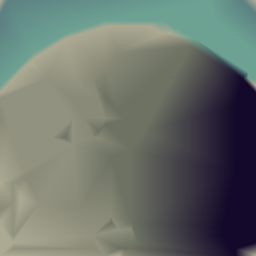} &\includegraphics[width=0.22\textwidth]{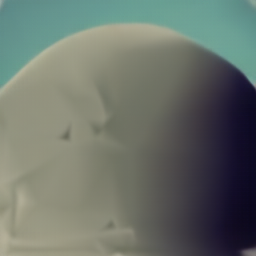} &\includegraphics[width=0.22\textwidth]{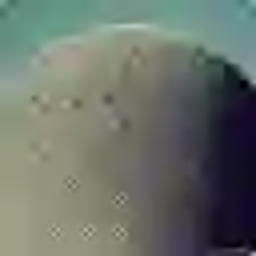} \\
\footnotesize{Ground Truth} & \footnotesize{13.2179 PSNR, 0.2111 SSIM} & \footnotesize{\textbf{13.2259 PSNR, 0.2121 SSIM}} & \footnotesize{13.2145 PSNR, 0.2086 SSIM} \\

\includegraphics[width=0.22\textwidth]{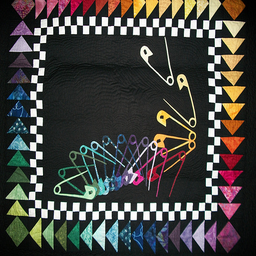} &\includegraphics[width=0.22\textwidth]{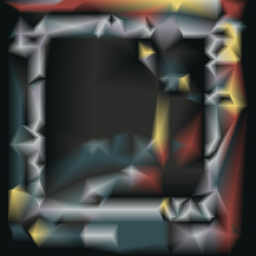} &\includegraphics[width=0.22\textwidth]{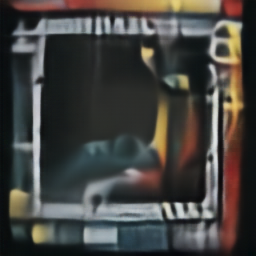} &\includegraphics[width=0.22\textwidth]{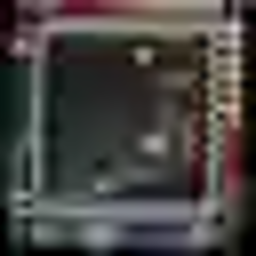}  \\
\footnotesize{Ground Truth} & \footnotesize{13.2704 PSNR, 0.3395 SSIM} & \footnotesize{\textbf{13.3209 PSNR, 0.3734 SSIM}} &\footnotesize{12.8824 PSNR, 0.2858 SSIM} \\

\includegraphics[width=0.22\textwidth]{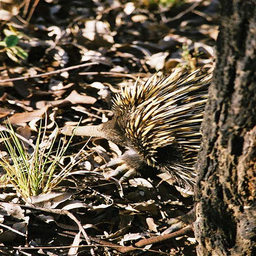} &\includegraphics[width=0.22\textwidth]{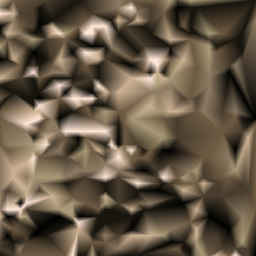} &\includegraphics[width=0.22\textwidth]{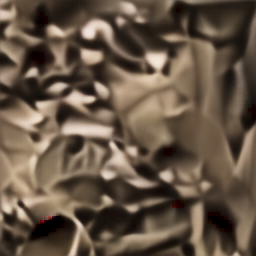} &\includegraphics[width=0.22\textwidth]{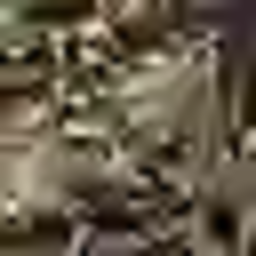} \\
\footnotesize{Ground Truth} & \footnotesize{13.0367 PSNR, 0.1830 SSIM} & \footnotesize{\textbf{13.4482 PSNR, 0.2189 SSIM}} & \footnotesize{12.6962 PSNR, 0.1356 SSIM}\\

\includegraphics[width=0.22\textwidth]{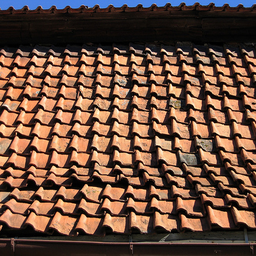} &\includegraphics[width=0.22\textwidth]{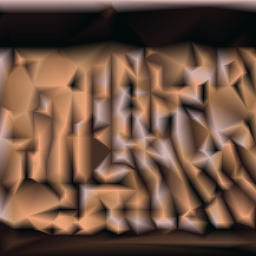} &\includegraphics[width=0.22\textwidth]{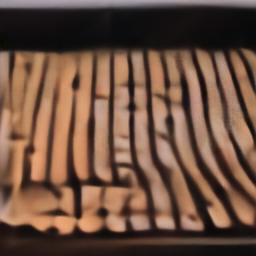} &\includegraphics[width=0.22\textwidth]{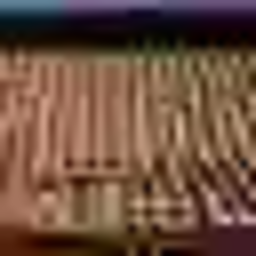} \\
\footnotesize{Ground Truth} & \footnotesize{12.7520 PSNR, 0.2868 SSIM} & \footnotesize{\textbf{13.5967 PSNR, 0.3586 SSIM}} & \footnotesize{12.4848 PSNR, 0.2032 SSIM} \\

  \end{tabular}
\end{figure*}

\begin{figure*}[h]
  \vspace*{-0.1in}
  \caption{Worst PSNR}
  \vspace*{-0.1in}
  \centering
  \begin{tabular}{cccc}
  \thead{Original Image} & \thead{Interpolated} & \thead{Neural Decoded} & \thead{WebP} \\
\includegraphics[width=0.22\textwidth]{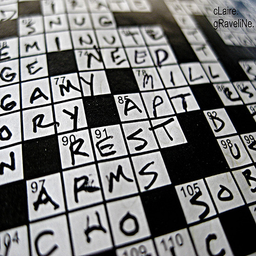} &\includegraphics[width=0.22\textwidth]{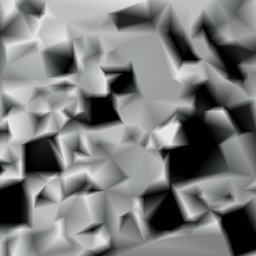} &\includegraphics[width=0.22\textwidth]{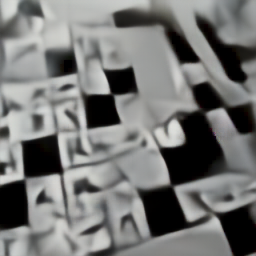}&
\includegraphics[width=0.22\textwidth]{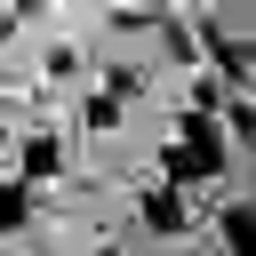}\\

\footnotesize{Ground Truth} & \footnotesize{12.0996 PSNR, 0.2213 SSIM} & \footnotesize{\textbf{12.8032 PSNR, 0.3029 SSIM}} & \footnotesize{11.6907 PSNR, 0.1503 SSIM} \\

\includegraphics[width=0.22\textwidth]{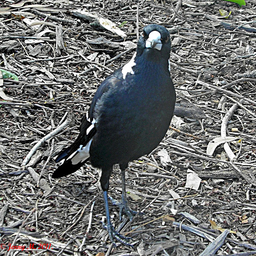} &\includegraphics[width=0.22\textwidth]{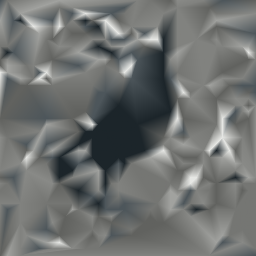} &\includegraphics[width=0.22\textwidth]{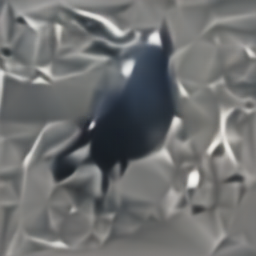}  &\includegraphics[width=0.22\textwidth]{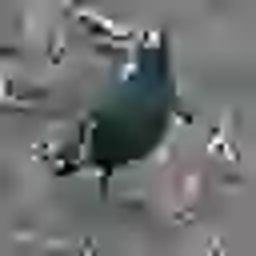}\\
\footnotesize{Ground Truth} & \footnotesize{13.1227 PSNR, 0.1299 SSIM} & \footnotesize{\textbf{13.2882 PSNR, 0.1496 SSIM}} & \footnotesize{13.0452 PSNR, 0.1211 SSIM} \\

\includegraphics[width=0.22\textwidth]{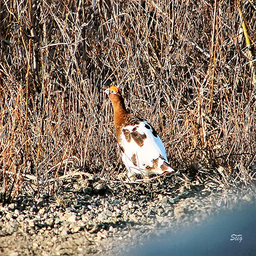} &\includegraphics[width=0.22\textwidth]{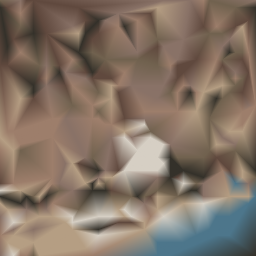} &\includegraphics[width=0.22\textwidth]{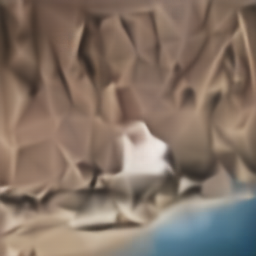}  &\includegraphics[width=0.22\textwidth]{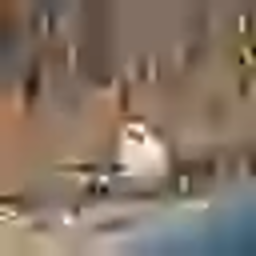}\\
\footnotesize{Ground Truth} & \footnotesize{13.2473 PSNR, 0.1040 SSIM} & \footnotesize{\textbf{13.4038 PSNR, 0.1157 SSIM}} & \footnotesize{13.1843 PSNR, 0.0973 SSIM} \\

\includegraphics[width=0.22\textwidth]{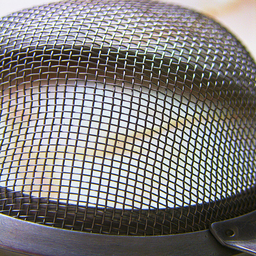} &\includegraphics[width=0.22\textwidth]{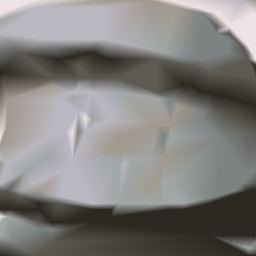} &\includegraphics[width=0.22\textwidth]{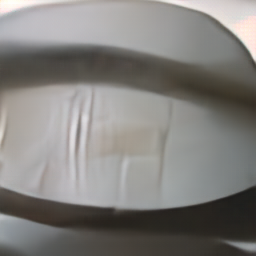}  &\includegraphics[width=0.22\textwidth]{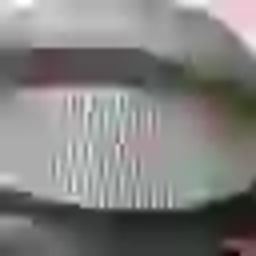}\\
\footnotesize{Ground Truth} & \footnotesize{13.3954 PSNR, 0.1427 SSIM} & \footnotesize{\textbf{13.4994 PSNR, 0.1541 SSIM}} & \footnotesize{13.4025 PSNR, 0.1437 SSIM} \\

\includegraphics[width=0.22\textwidth]{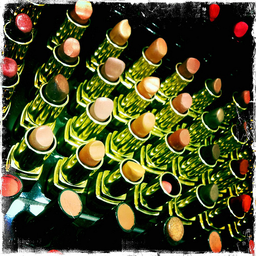} &\includegraphics[width=0.22\textwidth]{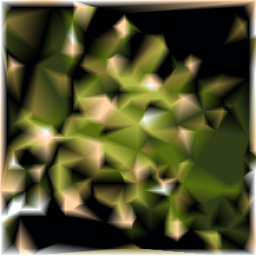} &\includegraphics[width=0.22\textwidth]{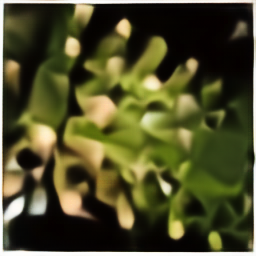} &\includegraphics[width=0.22\textwidth]{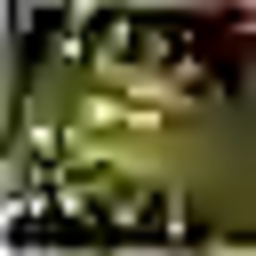}\\
\footnotesize{Ground Truth} & \footnotesize{13.1887 PSNR, 0.2672 SSIM} & \footnotesize{\textbf{13.6563 PSNR, 0.3064 SSIM}} & \footnotesize{11.8838 PSNR, 0.1647 SSIM}  \\

  \end{tabular}
\end{figure*}

\begin{figure*}[h]
  \vspace*{-0.1in}
  \caption{Best SSIM}
  \vspace*{-0.1in}
  \centering
  \begin{tabular}{cccc}
  \thead{Original Image} & \thead{Interpolated} & \thead{Neural Decoded} & \thead{WebP} \\
\includegraphics[width=0.22\textwidth]{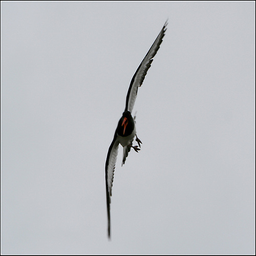} &\includegraphics[width=0.22\textwidth]{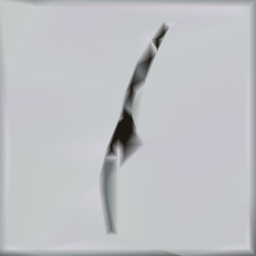} &\includegraphics[width=0.22\textwidth]{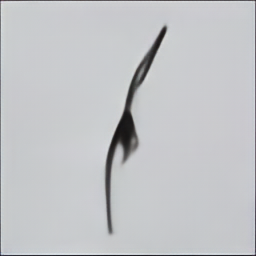} &\includegraphics[width=0.22\textwidth]{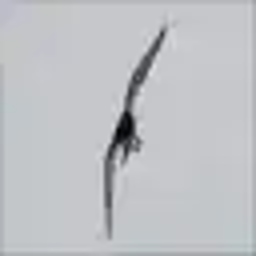}\\

\footnotesize{Ground Truth} & \footnotesize{23.0814 PSNR, 0.9077 SSIM} & \footnotesize{\textbf{27.0714 PSNR, 0.9474 SSIM}} & \footnotesize{23.4490 PSNR, 0.9231 SSIM}  \\

\includegraphics[width=0.22\textwidth]{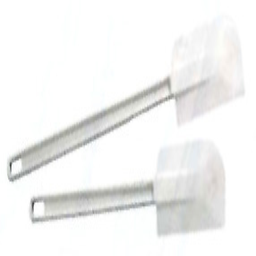} &\includegraphics[width=0.22\textwidth]{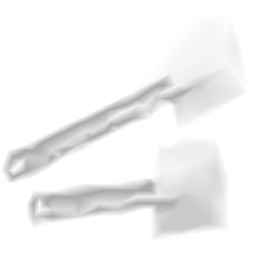} &\includegraphics[width=0.22\textwidth]{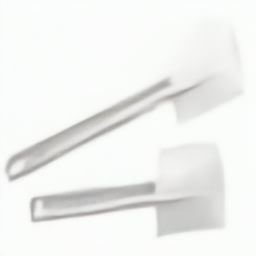} &\includegraphics[width=0.22\textwidth]{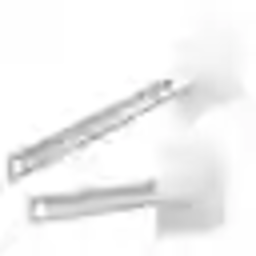}\\
\footnotesize{Ground Truth} & \footnotesize{30.5001 PSNR, 0.9176 SSIM} & \footnotesize{\textbf{32.3672 PSNR, 0.9495 SSIM}} &\footnotesize{30.8484 PSNR, 0.9180 SSIM} \\

\includegraphics[width=0.22\textwidth]{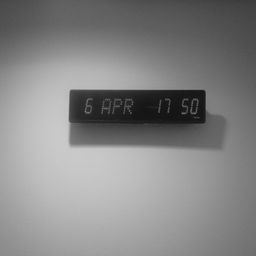} &\includegraphics[width=0.22\textwidth]{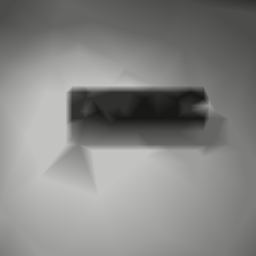} &\includegraphics[width=0.22\textwidth]{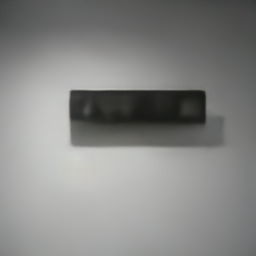} &\includegraphics[width=0.22\textwidth]{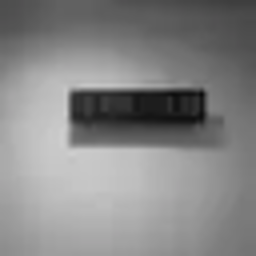}\\
\footnotesize{Ground Truth} & \footnotesize{29.7310 PSNR, 0.9296 SSIM} & \footnotesize{\textbf{35.3078 PSNR, 0.9522 SSIM}} & \footnotesize{33.9794 PSNR, 0.9413 SSIM} \\

\includegraphics[width=0.22\textwidth]{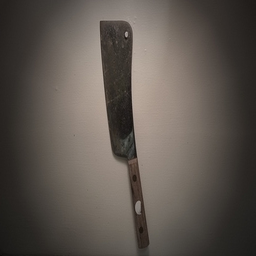} &\includegraphics[width=0.22\textwidth]{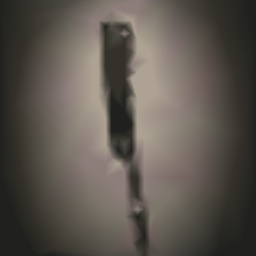} &\includegraphics[width=0.22\textwidth]{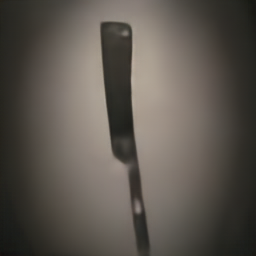} &\includegraphics[width=0.22\textwidth]{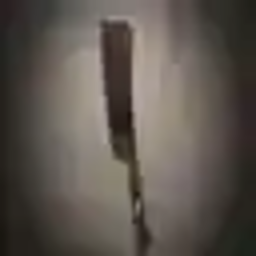}\\
\footnotesize{Ground Truth} & \footnotesize{30.6643 PSNR, 0.9273 SSIM} & \footnotesize{\textbf{34.9344 PSNR, 0.9540 SSIM}} & \footnotesize{31.6522 PSNR, 0.9306 SSIM} \\

\includegraphics[width=0.22\textwidth]{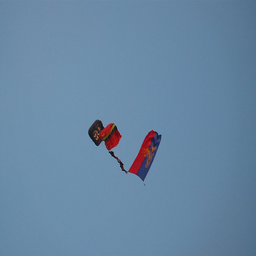} &\includegraphics[width=0.22\textwidth]{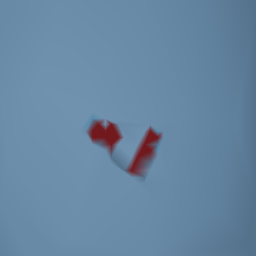} &\includegraphics[width=0.22\textwidth]{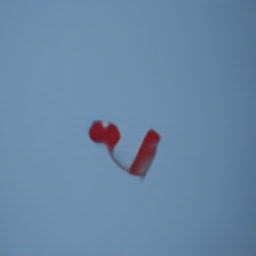} &\includegraphics[width=0.22\textwidth]{cLeAnEd_figure_130.png} \\
\footnotesize{Ground Truth} & \footnotesize{31.1331 PSNR, \textbf{0.9571 SSIM}} & \footnotesize{\textbf{31.7848 PSNR}, 0.9567 SSIM} & \footnotesize{31.6693 PSNR, 0.9548 SSIM} \\

  \end{tabular}
\end{figure*}

\begin{figure*}[h]
  \vspace*{-0.1in}
  \caption{Best SSIM}
  \vspace*{-0.1in}
  \centering
  \begin{tabular}{cccc}
  \thead{Original Image} & \thead{Interpolated} & \thead{Neural Decoded} & \thead{WebP} \\
\includegraphics[width=0.22\textwidth]{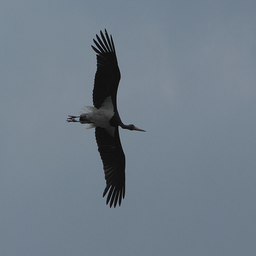} &\includegraphics[width=0.22\textwidth]{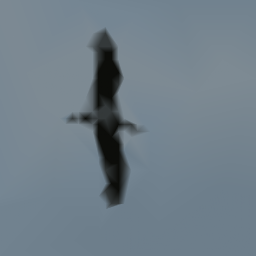} &\includegraphics[width=0.22\textwidth]{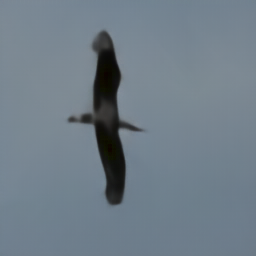} &\includegraphics[width=0.22\textwidth]{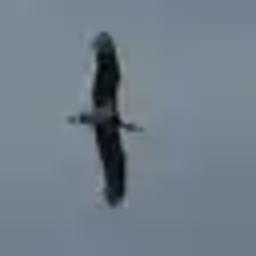}\\

\footnotesize{Ground Truth} & \footnotesize{29.5948 PSNR, 0.9361 SSIM} & \footnotesize{\textbf{31.0210 PSNR, 0.9480 SSIM}} & \footnotesize{30.1475, 0.9384 SSIM} \\

\includegraphics[width=0.22\textwidth]{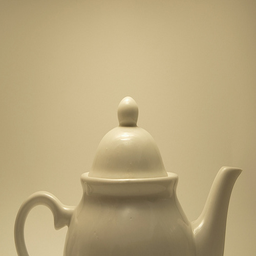} &\includegraphics[width=0.22\textwidth]{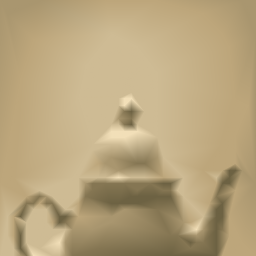} &\includegraphics[width=0.22\textwidth]{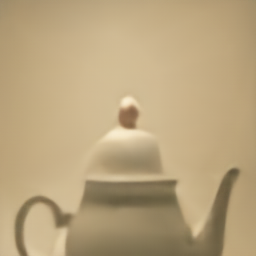} &\includegraphics[width=0.22\textwidth]{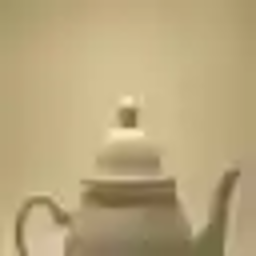} \\
\footnotesize{Ground Truth} & \footnotesize{31.7642 PSNR, 0.9299 SSIM} & \footnotesize{\textbf{34.0957 PSNR, 0.9505 SSIM}} & \footnotesize{32.7695 PSNR, 0.9360 SSIM} \\

\includegraphics[width=0.22\textwidth]{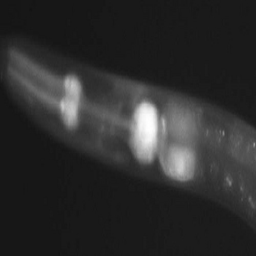} &\includegraphics[width=0.22\textwidth]{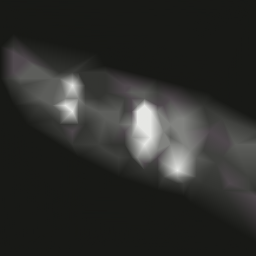} &\includegraphics[width=0.22\textwidth]{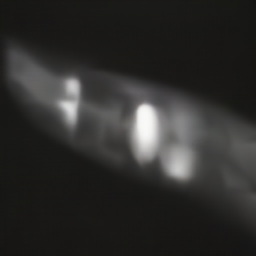} &\includegraphics[width=0.22\textwidth]{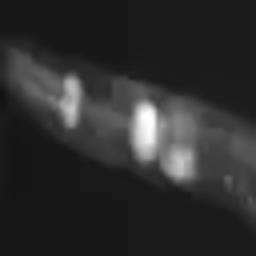}\\
\footnotesize{Ground Truth} & \footnotesize{32.7623 PSNR, 0.9441 SSIM} & \footnotesize{34.8865 PSNR, 0.9535 SSIM} & \footnotesize{\textbf{37.6550 PSNR, 0.9625 SSIM}} \\

\includegraphics[width=0.22\textwidth]{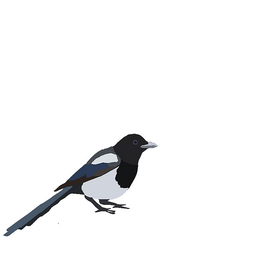} &\includegraphics[width=0.22\textwidth]{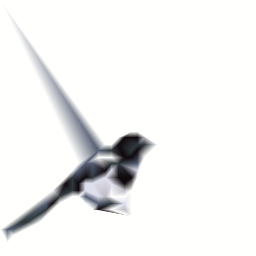} &\includegraphics[width=0.22\textwidth]{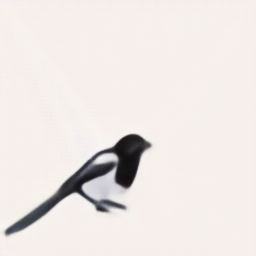} &\includegraphics[width=0.22\textwidth]{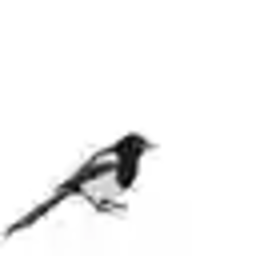}\\
\footnotesize{Ground Truth} & \footnotesize{21.4695 PSNR, 0.8972 SSIM} & \footnotesize{22.9038 PSNR, \textbf{0.9565 SSIM}} & \footnotesize{\textbf{25.5479 PSNR}, 0.9398 SSIM} \\

\includegraphics[width=0.22\textwidth]{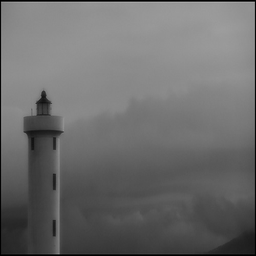} &\includegraphics[width=0.22\textwidth]{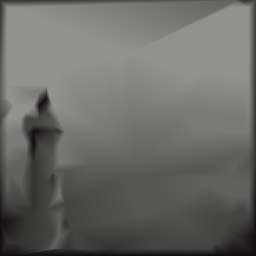} &\includegraphics[width=0.22\textwidth]{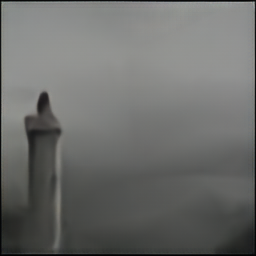} &\includegraphics[width=0.22\textwidth]{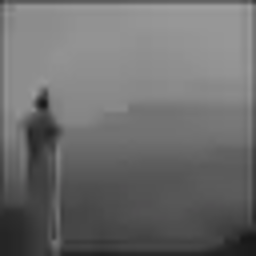}\\
\footnotesize{Ground Truth} & \footnotesize{25.5729 PSNR, 0.9287 SSIM} & \footnotesize{\textbf{33.2967 PSNR, 0.9612 SSIM}} & \footnotesize{26.4886 PSNR, 0.9374 SSIM} \\

  \end{tabular}
\end{figure*}

\begin{figure*}[h]
  \vspace*{-0.1in}
  \caption{Worst SSIM}
  \vspace*{-0.1in}
  \centering
  \begin{tabular}{cccc}
  \thead{Original Image} & \thead{Interpolated} & \thead{Neural Decoded} & \thead{WebP}\\
\includegraphics[width=0.22\textwidth]{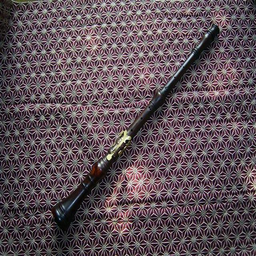} &\includegraphics[width=0.22\textwidth]{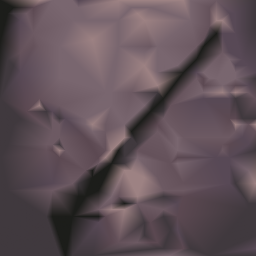} &\includegraphics[width=0.22\textwidth]{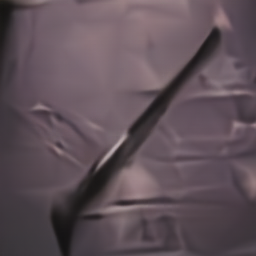} &\includegraphics[width=0.22\textwidth]{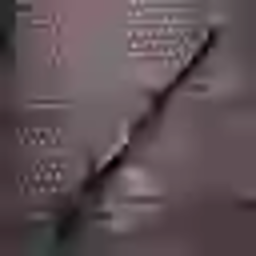}\\

\footnotesize{Ground Truth} & \footnotesize{15.5495 PSNR, 0.0659 SSIM} & \footnotesize{\textbf{15.6083 PSNR}, 0.0734 SSIM} &\footnotesize{15.5354 PSNR, \textbf{0.0735 SSIM}} \\

\includegraphics[width=0.22\textwidth]{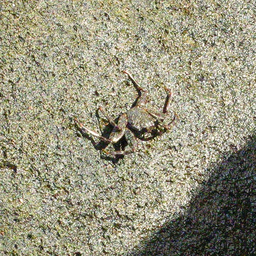} &\includegraphics[width=0.22\textwidth]{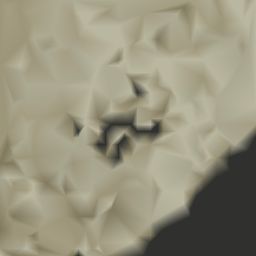} &\includegraphics[width=0.22\textwidth]{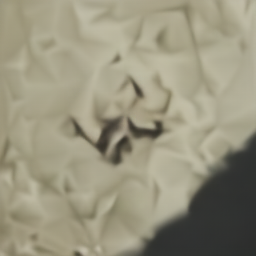} &\includegraphics[width=0.22\textwidth]{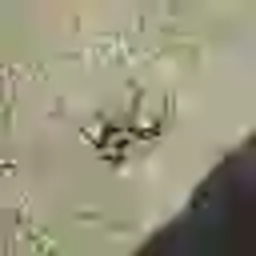}\\
\footnotesize{Ground Truth} & \footnotesize{16.5936 PSNR, 0.0833 SSIM} & \footnotesize{\textbf{16.7179 PSNR, 0.0909 SSIM}} & \footnotesize{16.5516 PSNR, 0.0860 SSIM} \\

\includegraphics[width=0.22\textwidth]{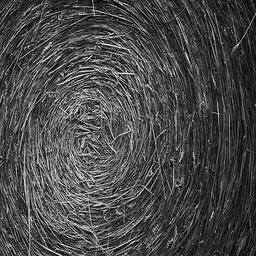} &\includegraphics[width=0.22\textwidth]{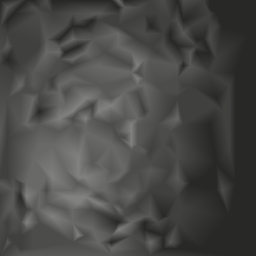} &\includegraphics[width=0.22\textwidth]{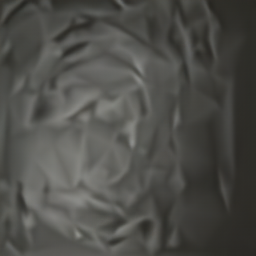} &\includegraphics[width=0.22\textwidth]{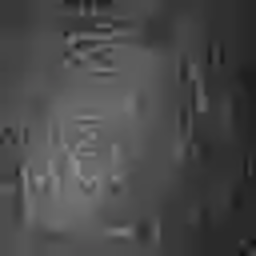}\\
\footnotesize{Ground Truth} & \footnotesize{16.4997 PSNR, 0.0879 SSIM} & \footnotesize{\textbf{16.5101 PSNR, 0.0944 SSIM}} & \footnotesize{16.4033 PSNR, 0.08411 SSIM}\\

\includegraphics[width=0.22\textwidth]{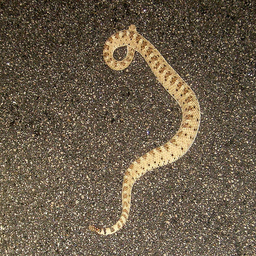} &\includegraphics[width=0.22\textwidth]{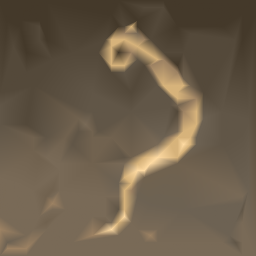} &\includegraphics[width=0.22\textwidth]{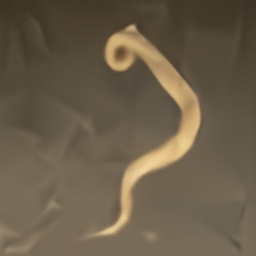} &\includegraphics[width=0.22\textwidth]{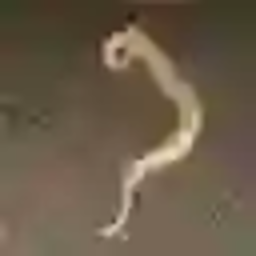} \\
\footnotesize{Ground Truth} & \footnotesize{18.2631 PSNR, 0.0962 SSIM} & \footnotesize{\textbf{18.3455 PSNR, 0.1022 SSIM}} & \footnotesize{18.0795 PSNR, 0.0917 SSIM} \\

\includegraphics[width=0.22\textwidth]{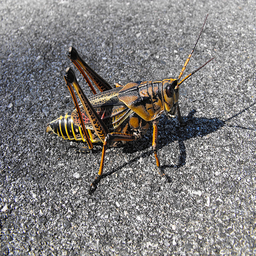} &\includegraphics[width=0.22\textwidth]{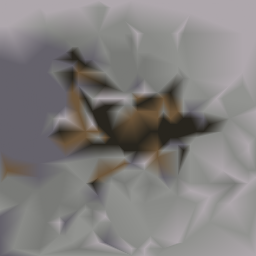} &\includegraphics[width=0.22\textwidth]{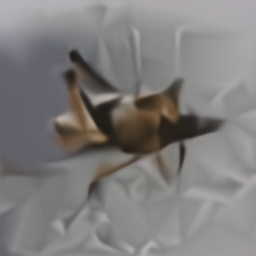} &\includegraphics[width=0.22\textwidth]{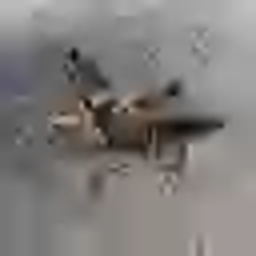}\\
\footnotesize{Ground Truth} & \footnotesize{16.6109 PSNR, 0.1014 SSIM} & \footnotesize{\textbf{16.7528 PSNR, 0.1111 SSIM}} & \footnotesize{16.4989 PSNR, 0.0962 SSIM} \\

  \end{tabular}
  \caption{Worst SSIM}
\end{figure*}

\begin{figure*}[h]
  \vspace*{-0.1in}
  \caption{Worst SSIM}
  \vspace*{-0.1in}
  \centering
  \begin{tabular}{cccc}
  \thead{Original Image} & \thead{Interpolated} & \thead{Neural Decoded} & \thead{WebP}\\
\includegraphics[width=0.22\textwidth]{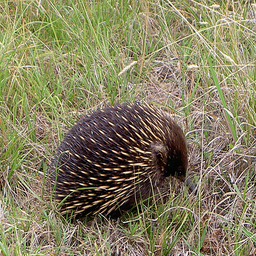} &\includegraphics[width=0.22\textwidth]{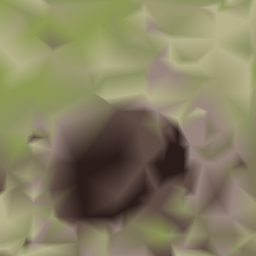} &\includegraphics[width=0.22\textwidth]{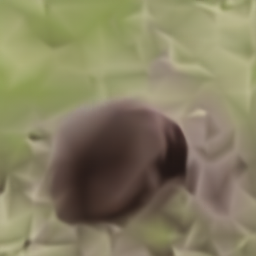} &\includegraphics[width=0.22\textwidth]{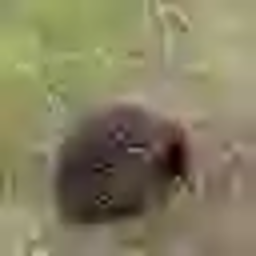}\\

\footnotesize{Ground Truth} & \footnotesize{16.0347 PSNR, 0.0777 SSIM} & \footnotesize{\textbf{16.1087 PSNR, 0.0814 SSIM}} & \footnotesize{15.9726 PSNR, 0.0765 SSIM} \\

\includegraphics[width=0.22\textwidth]{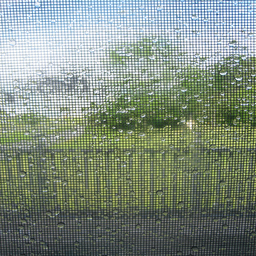} &\includegraphics[width=0.22\textwidth]{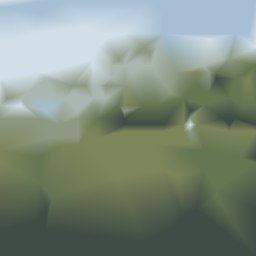} &\includegraphics[width=0.22\textwidth]{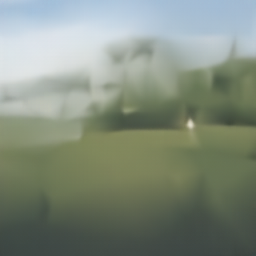} &\includegraphics[width=0.22\textwidth]{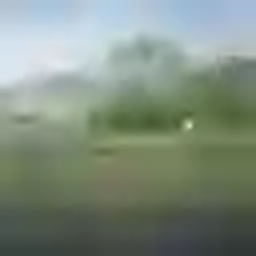}\\
\footnotesize{Ground Truth} & \footnotesize{17.7205 PSNR, 0.0933 SSIM} & \footnotesize{\textbf{17.7716 PSNR, 0.0935 SSIM}} & \footnotesize{17.7348 PSNR, 0.0909 SSIM} \\

\includegraphics[width=0.22\textwidth]{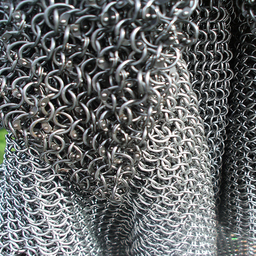} &\includegraphics[width=0.22\textwidth]{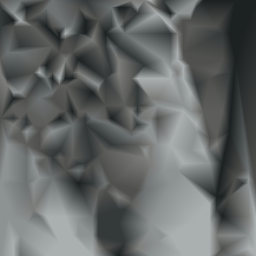} &\includegraphics[width=0.22\textwidth]{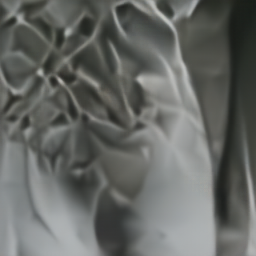} &\includegraphics[width=0.22\textwidth]{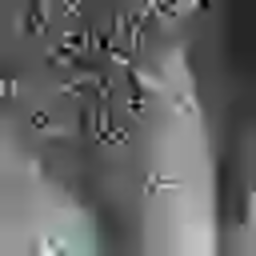}\\
\footnotesize{Ground Truth} & \footnotesize{13.4859 PSNR, 0.0805 SSIM} & \footnotesize{\textbf{13.6928 PSNR, 0.1000 SSIM}} &\footnotesize{13.4264 PSNR, 0.0754 SSIM} \\

\includegraphics[width=0.22\textwidth]{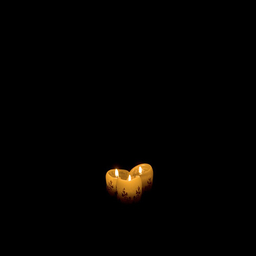} &\includegraphics[width=0.22\textwidth]{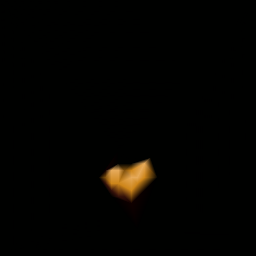} &\includegraphics[width=0.22\textwidth]{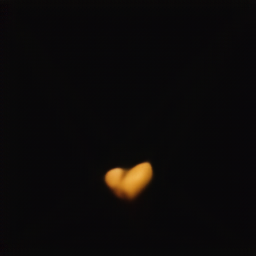} &\includegraphics[width=0.22\textwidth]{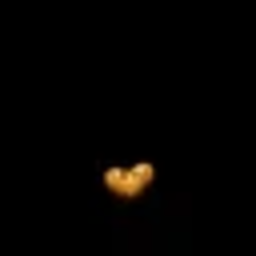}\\
\footnotesize{Ground Truth} & \footnotesize{33.1259 PSNR, 0.9421 SSIM} & \footnotesize{28.6035 PSNR, 0.1032 SSIM} & \footnotesize{\textbf{34.9583 PSNR, 0.9679 SSIM}} \\ 

\includegraphics[width=0.22\textwidth]{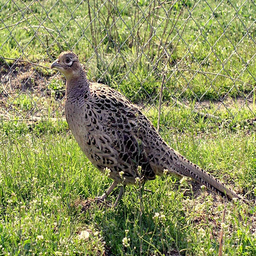} &\includegraphics[width=0.22\textwidth]{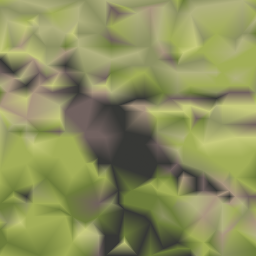} &\includegraphics[width=0.22\textwidth]{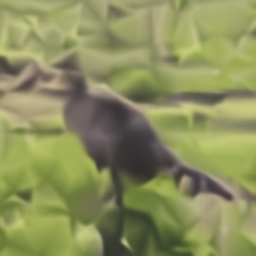} &\includegraphics[width=0.22\textwidth]{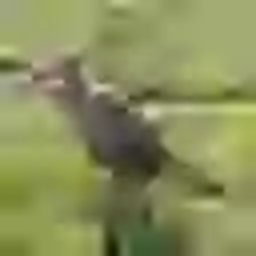}\\
\footnotesize{Ground Truth} & \footnotesize{16.5107 PSNR, 0.1072 SSIM} & \footnotesize{\textbf{16.6494 PSNR, 0.1148 SSIM}} & \footnotesize{16.0610 PSNR, 0.0815 SSIM} \\

  \end{tabular}
\end{figure*}

\end{document}